\documentclass[journal]{IEEEtran}

\usepackage[utf8]{inputenc}
\usepackage[T1]{fontenc}
\usepackage{siunitx}
\usepackage[colorlinks=true, allcolors=blue]{hyperref}
\usepackage{amsmath}
\usepackage{amsfonts}
\usepackage{amssymb}
\usepackage{gensymb}
\usepackage{booktabs}
\usepackage{soul} %ONLY FOR THE DRAFT VERSION, REMOVE AFTERWARDS
\usepackage{subcaption}
\usepackage{stfloats}
\usepackage{booktabs}
\usepackage{enumitem}
\usepackage[justification=centering]{caption}
\usepackage{tikz,xcolor}
\usepackage{makecell}

\definecolor{lime}{HTML}{A6CE39}
\DeclareRobustCommand{\orcidicon}
{
    \begin{tikzpicture}
    \draw[lime, fill=lime] (0,0) circle [radius=0.16] 
    node[white] {{\fontfamily{qag}\selectfont \tiny ID}};    \draw[white, fill=white] (-0.0625,0.095) circle [radius=0.007];    
    \end{tikzpicture}
    \hspace{0mm}}
\foreach \x in {A, ..., Z}{%
    \expandafter\xdef\csname orcid\x\endcsname{\noexpand\href{https://orcid.org/\csname orcidauthor\x\endcsname}{\noexpand\orcidicon}}
}

\renewcommand{\tablename}{Table}

\setlist[enumerate]{itemsep = 0pt, parsep = 0pt, topsep = 0pt} % enumerate spacing setting
\setlist[itemize]{itemsep = 0pt, parsep = 0pt, topsep = 0pt} % itemize spacing setting

\begin{document}

\title{Automatic parking planning control method based on improved A* algorithm}

% https://www.programmersought.com/article/12636730965/
\author{Yuxuan Zhao
% \thanks{This work was supported in part by the National Natural Science Foundation of China under Grant 62261160654, in part by the Shenzhen Fundamental Research Program under Grant JCYJ20220818103006012, in part by GuangDong Basic and Applied Basic Research Foundation under Grant 2021A1515110641, in part by the Shenzhen Key Laboratory of Robotics and Computer Vision (ZDSYS20220330160557001), in part by the Science and Technology Development Fund of Macao S.A.R (FDCT) under Grant No. 0123/2022/AFJ and Grant No. 0081/2022/A2. (Corresponding authors: Gongjin Lan; Qi Hao.)}
% \thanks{Gongjin Lan, XXX are with the Department of Computer Science and Engineering, Southern University of Science and Technology, Shenzhen, 518055, China (e-mail: langj@sustech.edu.cn)}
% \thanks{Qi Hao is with the Department of Computer Science and Engineering, Research Institute of Trustworthy Autonomous Systems, Southern University of Science and Technology, Shenzhen, 518055, China (e-mail: hao.q@sustech.edu.cn)}
% \thanks{Fei Hu is with the Department of Electrical and Computer Engineering, University of Alabama, Tuscaloosa, AL (email:fei@eng.ua.edu)}
}

% The paper headers
\markboth{Journal of \LaTeX\ Class Files,~Vol.
% ~14, No.~8, August~2015
}%
{Shell \MakeLowercase{\textit{et al.}}: Bare Demo of IEEEtran.cls for IEEE Journals}

% make the title area
\maketitle

% As a general rule, do not put math, special symbols or citations
% in the abstract or keywords.
\begin{abstract}

As the trend of moving away from high-precision maps gradually emerges in the autonomous driving industry, traditional planning algorithms are gradually exposing some problems.
To address the high real-time, high precision, and high trajectory quality requirements posed by the automatic parking task under real-time perceived local maps, this paper proposes an improved automatic parking planning algorithm based on the A* algorithm, and uses Model Predictive Control (MPC) as the control module for automatic parking.
The algorithm enhances the planning real-time performance by optimizing heuristic functions, binary heap optimization, and bidirectional search; it calculates the passability of narrow areas by dynamically loading obstacles and introduces the vehicle's own volume during planning; it improves trajectory quality by using neighborhood expansion and Bezier curve optimization methods to meet the high trajectory quality requirements of the parking task.
After obtaining the output results of the planning algorithm, a loss function is designed according to the characteristics of the automatic parking task under local maps, and the MPC algorithm is used to output control commands to drive the car along the planned trajectory.
This paper uses the perception results of real driving environments converted into maps as planning inputs to conduct simulation tests and ablation experiments on the algorithm.
Experimental results show that the improved algorithm proposed in this paper can effectively meet the special requirements of automatic parking under local maps and complete the automatic parking planning and control tasks.

\end{abstract}

% Note that keywords are not normally used for peerreview papers.

% For peer review papers, you can put extra information on the cover
% page as needed:
% \ifCLASSOPTIONpeerreview
% \begin{center} \bfseries EDICS Category: 3-BBND \end{center}
% \fi
%
% For peerreview papers, this IEEEtran command inserts a page break and
% creates the second title. It will be ignored for other modes.
\IEEEpeerreviewmaketitle

%%%%%%%%%%%%%%%%%%%%%%%%%%%%%%%%%%%%%%%%%%%%%%
\section{Introduction}
%%%%%%%%%%%%%%%%%%%%%%%%%%%%%%%%%%%%%%%%%%%%%%

\subsection{Research Background and Significance}
\label{Research Background and Significance}

% Since 2023, automobiles have become one of the main drivers of China's economic growth. 
Autonomous driving is currently a hot topic in the automotive industry and has become a key area of intense competition among numerous domestic and international manufacturers and intelligent driving solution providers \cite{lan2023virtual,yi2024key}. 
Once vehicles equipped with advanced autonomous driving become widespread, they will completely overturn people's travel choices.
Tech companies such as Huawei, Tesla, Google, Baidu, and Xiaomi, as well as research institutes like Tsinghua University and the Shanghai Institute of Artificial Intelligence, have entered the field of autonomous driving, continuously launching products equipped with intelligent driving features.
In recent years, research institutes have published a large number of high-quality papers.
% As early as 2020, China's National Development and Reform Commission and Ministry of Industry and Information Technology issued the "Intelligent Vehicle Innovation and Development Strategy" \cite{ref1}.
% This document proposes requirements for "intelligence," "networking," and "standardization," demanding that China's autonomous driving vehicle technology innovation, industrial ecology, infrastructure, and the regulatory standard system should be basically formed by 2025.
% In July 2021, China's Ministry of Industry and Information Technology released opinions on the access management of intelligent connected vehicles, laying the policy foundation for the R\&D and mass production of high-level intelligent driving vehicles.
It can be said that from the government to the private sector, from domestic to international, autonomous driving vehicles have become a core topic in the development of technology and the economy.

Automatic parking is a very important part of the autonomous driving scenario. The automatic parking system can liberate drivers, save a lot of time and energy, and optimize passengers' driving experience.
Moreover, parking is a part of the driving process with a high incidence of accidents, and the large-scale application of high-performance automatic parking systems can significantly reduce traffic accidents during parking.
With the continuous development of automatic parking systems and the further popularization of intelligent cars supporting automatic parking, building unmanned automatic parking lots with small footprints and high efficiency will become a reality from theory.

In recent years, revolutionary new technologies for the perception end of autonomous driving have emerged one after another, and the wave of moving away from high-precision maps has intensified. 
The joint perception technology of Bird's Eye View (BEV) \cite{ref5} + Transformer \cite{ref6} + Occupancy \cite{ref10} has greatly improved the perception accuracy of intelligent cars for the environment, making it possible to abandon prior information of high-precision maps and save a lot of map surveying and administrative costs. 
Autonomous driving vehicles will have the ability to obtain precise information about the surrounding environment without relying on pre-acquired high-precision maps, solely depending on the sensors carried by the vehicle itself \cite{xu2019online}. 
In this way, autonomous driving vehicles can normally activate intelligent driving functions in areas without high-precision maps, greatly expanding the usage scenarios of autonomous driving.

\subsection{Overseas and Domestic Research Status}
\label{Development Status at Home and Abroad}

The origin of the automatic parking system can be traced back to the IRVW Futura concept car launched by Volkswagen in 1992, but it was not put into production due to its higher cost than the market's ability to accept. 
It was not until 2003 that Toyota pioneered the commercialization of automatic parking on the Prius.
In 2004, the Evolve project developed by Linkoping University in Sweden, collaborating with Volvo, also demonstrated automatic parking technology. 
These advancements marked the initial application of automatic parking technology. 
As of 2020, the installation rate of Autonomous Parking System (APS) in the domestic passenger car market has reached 12.3\%.
In the foreseeable future, the installation rate of automatic parking systems will grow rapidly, reflecting significant market potential.

China entered the field of autonomous driving and automatic parking later, but its enthusiasm for development is high. 
The central and local governments actively encourage pilot projects, promoting regulatory improvements and policy relaxation.
Chinese consumers are enthusiastic about autonomous driving, embracing the electrification and intelligence of automobiles. 
With the improvement of the industrial chain, a large number of enterprises are emerging to build high-level autonomous driving and automatic parking application scenarios.

In recent years, domestic and foreign scholars have conducted numerous studies on various components of automatic parking systems, including parking space detection, path planning, tracking control, and more.
Existing planning methods include graph-based search methods, random sampling-based methods, spline curve-based methods, numerical optimization-based methods, and the more recently emerging machine learning-based methods. 
The specific implementation principles, advantages, and disadvantages of these planning methods will be elaborated in detail in \autoref{Traditional Planning Methods}.

\subsection{Main Contributions}
\label{Main Contributions}

\begin{enumerate}
    \item Optimize the A* algorithm to enhance its planning speed, meeting the high real-time requirements of local map parking planning.
    \item Optimize the A* algorithm to improve its trajectory quality, satisfying the requirements of high precision and high trajectory quality for automatic parking tasks.
    \item Introduce the volume of the ego-vehicle into the planning process to better simulate real parking scenarios.
    \item Implement the MPC control algorithm for targeted nonlinear optimization of automatic parking paths.
    \item Conduct simulation tests and ablation experiments on the algorithms using Python.
\end{enumerate}

%%%%%%%%%%%%%%%%%%%%%%%%%%%%%%%%%%%%%%%%%%%%%%
\section{Related Work}
\label{sec:related_work}
%%%%%%%%%%%%%%%%%%%%%%%%%%%%%%%%%%%%%%%%%%%%%%

\subsection{Characteristics of Autonomous Driving Parking Tasks}
\label{Characteristics of Autonomous Driving Parking Tasks}

Automated Valet Parking (AVP), as an important subdivision of autonomous driving planning modules, has garnered significant attention in recent years.
By 2024, automatic parking has become a hot topic in the field of autonomous driving and a focal point of research and development for various companies in the industry.
Automatic parking in underground parking garages accounts for a large proportion of all automatic parking tasks.
Underground garage scenarios typically exhibit the following characteristics: dim lighting and severe obstruction of vision by a large number of obstacles, including pillars \cite{ref14,lan2022vision}. 
Complex obstacles can increase vehicle positioning errors and planning difficulties. 
In such scenarios, traditional methods inevitably suffer from significant interference, resulting in unsatisfactory planning results.

Both domestic and international researchers widely consider the parking scenario a highly challenging driving task. 
Vehicles must complete a series of actions such as reversing and fine steering in the narrow space near the parking space, which also greatly increases the risk of collision \cite{ref15}. 
To address these challenges, planning and control algorithms need to plan appropriate trajectories to complete the action of parking in a parking space, as well as finely control the displacement of the vehicle in narrow spaces to avoid obstacles such as pillars, other vehicles, and pedestrians \cite{lan2023end}. 
The complexity of the parking scenario requires high robustness of the parking planning algorithm.

The underground garage scenario poses stringent requirements on planning algorithms.
During the parking process, there are many uncontrollable factors, such as changes in the steering system's execution speed and accuracy, and changes in obstacle positions, which may lead to deviations between the actual parking trajectory and the planned trajectory.
Therefore, planning algorithms need to have the ability to dynamically correct trajectories in real-time or even re-plan.
Current parking scenarios are mainly divided into vertical parking and parallel parking, and the planning algorithm also needs to support these two main parking scenarios.
Simultaneously, the control algorithm also needs to ensure that the vehicle is parked accurately and without deviation when entering the parking space. 
In narrow underground garage environments, even small differences in planning accuracy can lead to collisions, so high precision is an indispensable characteristic of the planning algorithm. 
Even if the planned path quality is high, if the control algorithm outputs low-quality control commands, leading to distorted or deviated actual driving trajectories, it can also result in dangerous accidents.
Therefore, trajectory quality and control accuracy are also important indicators that need to be considered in the planning and control algorithm.

Automatic parking planning methods need to accurately find high-quality paths in the cluttered environment of the underground garage; the control part also needs to minimize control errors.
A high-performance automatic parking system must take into account various requirements such as perception, positioning, motion planning, and actuator accuracy \cite{ref17}. 
Because of this, the automatic parking function is only available as an optional feature on mid-to-high-end models priced at 200,000 or above.
Therefore, the automatic parking solution proposed in this paper must consider the inherent complexity of the automatic parking task and the exceptionally high requirements for planning accuracy and trajectory quality.

In summary, automatic parking tasks under local maps need to face two major issues:
\begin{enumerate}
    \item Local map perception requires planning algorithms to have high real-time performance and fast computation speed.
    \item Automatic parking tasks require planning algorithms to generate trajectories with high quality, stable, and safe driving.
\end{enumerate}

\subsection{Traditional Planning Methods}
\label{Traditional Planning Methods}

Autonomous driving planning aims to enable vehicles to safely reach the target position from the starting point while satisfying global and local constraints \cite{ref18}.
In autonomous driving systems, grid maps are often used to represent the occupancy status of the vehicle's surrounding environment.
Currently, various algorithms have been applied to the development of autonomous vehicles, which can be broadly classified into graph search algorithms \cite{lan2022class,gao2021neat}, random sampling search algorithms, interpolation curve planning algorithms, and numerical optimization algorithms \cite{ref19,lan2022time,lan2021learning,lan2021learning2}.

The Dijkstra algorithm is a graph search algorithm used to find the shortest path from a single source in a graph.
It performs a breadth-first search from the starting point, examining the occupancy status of the vehicle and surrounding grids.
This algorithm is capable of finding the shortest path between any two points in the graph, ensuring that an optimal path from the current position of the self-driving vehicle to the planning target endpoint is generated when a feasible path exists. 
In the field of autonomous driving, the Dijkstra algorithm has been practically applied in the Darpa Urban Challenge \cite{ref21}.
The advantage of this algorithm is that it can generate the best planning path in a static semantic map with road information.

The A-star (A* algorithm) is a heuristic graph search algorithm that improves upon the Dijkstra algorithm.
The A* algorithm achieves more efficient search by defining weights for nodes. 
The A* algorithm performs well when global map information is available, but as the map size increases, its search cost grows exponentially, making it less effective in handling large-scale planning problems.

The state grid planning method discretizes continuous space into a state grid map, reducing the planning complexity in real space. 
In the state grid, the connection between each vertex is based on the vehicle's kinematic model, thus ensuring that all paths are practically feasible \cite{ref23}. 
This method incorporates kinematic constraints into the ordinary grid map, ensuring the continuity of states.

The Rapidly-exploring Random Tree (RRT) algorithm is a planning algorithm based on random sampling.
It takes the current position of the vehicle as the root node and generates a random expansion tree by randomly selecting reachable points as new leaf nodes, allowing the search tree to continuously grow towards the planning target endpoint, thus finding a feasible path. 
The RRT algorithm allows for rapid planning in semi-structured spaces, generating feasible solutions in a short time and converging asymptotically to optimal solutions \cite{ref25}.
However, as actual vehicles may not be able to move exactly according to the path required by the random search tree, it is often necessary to incorporate constraints such as steering angles and use interpolation methods to smooth the initial path in practical applications \cite{ref26}.

The Probabilistic Roadmap (PRM) algorithm is another planning algorithm based on sampling. 
It randomly generates a large number of nodes in the search space and connects them. 
Then, a collision-free path from the starting point to the endpoint is selected in the completed graph, forming a feasible trajectory.

Interpolation curve planning methods require predetermined nodes, and then use interpolation algorithms to generate navigation trajectories that satisfy trajectory continuity and vehicle kinematic constraints. 
The advantage of this method is that it can generate smooth curved paths while ensuring continuous and smooth velocity and acceleration \cite{ref27}, making it suitable for scenarios with high requirements for obstacle avoidance, path quality, and trajectory comfort. However, for non-standard parking spaces or complex environments, the effectiveness of interpolation curve planning methods may be limited \cite{ref28}.

The two-arc method and the arc-line combination method can be collectively referred to as geometric curve methods, which are concise and widely used parking path planning methods. 
They take two arcs as the basic path, directly connecting the two arcs or connecting them with a straight line \cite{ref30}.
However, this approach may result in curvature discontinuity points, making the planned path have discontinuous curvature and not satisfying the vehicle's kinematic constraints. 
Therefore, in practical applications, it is still necessary to combine other interpolation curve planning methods to smooth the curvature discontinuity points \cite{ref31}.

Trajectory planning for automatic parking utilizes various algorithms, including the clothoid curve planning that employs the Fresnel integral-defined clothoid curve to plan paths \cite{ref32}. 
This curve is capable of defining trajectories with linearly varying curvature, enabling a smooth transition between curved and straight sections of the trajectory. 
This method is easy to implement and generates paths with high approximation accuracy.

Polynomial planning methods satisfy longitudinal and lateral constraints using polynomials of different degrees, generating safe planned trajectories \cite{ref33}.
They excel in fitting position, angle, and curvature constraints.

Spline planning methods utilize segmented polynomial parameters and generate high-order continuously differentiable polynomials through linear combinations of spline basis functions to accomplish trajectory fitting.
This approach can generate highly smooth trajectories with good smoothness at the connection points of spline segments. 
Currently, spline planning methods such as cubic spline curves, Bezier curves, and B-spline curves \cite{ref34,ref36} have been widely applied in autonomous driving projects for various scenarios, including mining and passenger transportation \cite{ref38}.

Numerical optimization-based parking path planning methods seek optimal paths by minimizing or maximizing functions subject to constrained variables. 
However, this method is notably disadvantageous due to its long computation time and low efficiency, thus it has not been widely adopted for parking planning tasks.
It is typically used for smoothing initial trajectories or in scenarios with stringent kinematic constraints.

In addition to path planning algorithms, academia is also exploring the use of machine learning algorithms to achieve automatic parking functionality. 
For instance, a research group from Jiangsu University attempted to utilize BP neural networks to construct a driver model to simulate parking behavior.
Meanwhile, researchers have also proposed using genetic algorithms to train vehicles to park in simulation environments \cite{ref40,lan2019simulated}, but their achievements are only applicable to very simple standard parking spaces and are difficult to handle complex underground garage scenarios with multiple obstacles. 
Training neural networks is not only time-consuming but also computationally intensive.
More importantly, collecting a large amount of high-quality driving data for training is a significant challenge. 
Currently, publicly available high-quality datasets are relatively scarce. 
Although machine learning-based automatic parking controllers exhibit good performance in some aspects, they have not significantly surpassed traditional path planning-based automatic parking algorithms in terms of performance. 
Their generalization and robustness still need to be improved when facing complex environments such as underground garages.
Therefore, in developing automatic parking planning and control methods under local maps, machine learning-based parking methods do not need to be the preferred choice.

Common traditional autonomous driving planning methods and their respective advantages and disadvantages are presented in \autoref{tab:widgets}:

\begin{table*} \centering \small
\setlength\tabcolsep{2pt} \renewcommand{\arraystretch}{1.0}
\begin{tabular}{|l|l|l|} \toprule
\textbf{Method} & Advantages & Disadvantages \\ \hline
\textbf{Graph Search Algorithms} & & \\ \hline
Dijkstra \cite{ref21} & Ensures generation of optimal paths & Inefficient algorithm execution \\ \hline
A* & Heuristic search & Paths may not be continuous \\ \hline
State Grid Planning \cite{ref23} & Handles multi-dimensional information & High computational cost \\ \hline
\textbf{Random Sampling Algorithms} & Quickly generates local optimal solutions & Difficulty balancing optimality and real-time performance \\ \hline
RRT \cite{ref25,ref26} & Guarantees algorithm convergence & Generated trajectories may be discontinuous and uneven \\ \hline
PRM & & Long obstacle detection time \cite{ref44}\\ \hline
\textbf{Interpolation Curve Planning} & Generates paths with continuous curvature & Less effective with many obstacles \\ \hline
Arc-Straight Line \cite{ref30,ref31} & Simple calculation & Discontinuous curvature \\ \hline
Clothoid Curve Planning \cite{ref32} & Easy tracking control implementation & Many control nodes, complex calculations \\ \hline
Polynomial Planning \cite{ref33} & Low computational cost & Non-zero curvature at endpoints, requires large space \\ \hline
Spline Planning \cite{ref34,ref36,ref38} & Low computational cost & Difficulty ensuring optimality \\ \hline
\textbf{Numerical Optimization} & Meets complex constraint requirements & High computational load \\ \hline
\textbf{Artificial Intelligence} \cite{ref40} & & Insufficient quality data \\ \hline
\end{tabular}
\renewcommand{\tablename}{Table}
\caption{\label{tab:widgets}Advantages and Disadvantages of Traditional Planning Methods}
\end{table*}
 
The aforementioned planning methods have their own limitations, ranging from slow computation speed and unsatisfactory trajectory generation quality to the need for extensive high-quality data for training, which is costly and inefficient. 
These deficiencies hinder the algorithms from efficiently accomplishing underground garage parking tasks relying solely on existing methods.

In recent years, despite advancements in planning methods, there has been no revolutionary technological breakthrough. 
To further enhance the performance of autonomous driving planning and automatic parking systems and make them suitable for the specific scenario of underground garage parking, this study must explore new planning methods or improve existing ones.
Especially in the context of abandoning high-precision maps and adopting Bird's Eye View (BEV) local maps for planning, the limitations of traditional methods become more apparent.

The primary difference between local maps and global maps lies in their real-time requirements.
Global path planning is static, while local path planning is dynamic. 
Autonomous vehicles need to continuously perceive the surrounding environment during driving, generating rapidly updated local maps centered on themselves. 
This requires planning algorithms to process updated maps in a short time and quickly plan paths based on the new maps.

However, traditional planning methods often rely on prior map information.
When the environment changes or only local environmental information is available, they often cannot handle these changes promptly. 
This can result in excessive computation time and suboptimal planning results, which are unacceptable for automatic parking tasks requiring fine planning control.

Specifically, the A* algorithm based on graph search is still not fast enough; the RRT algorithm based on random sampling may fail to generate suboptimal or feasible paths when planning time is insufficient or sampling density is low, and the paths may be discontinuous or non-smooth; the PRM algorithm has a long computation time in environments with many obstacles like underground garages and performs poorly in narrow spaces; interpolation curve planning algorithms have rapidly increasing computation time in complex environments, making it difficult to plan feasible routes; while machine learning-based planning methods face the issue of a lack of high-quality driving data suitable for local map adaptation.

In summary, this study needs to integrate and improve various planning algorithms, targeting the real-time requirements of local maps and the demand for planning accuracy and trajectory quality in underground garage parking tasks. 
This approach can better address the complex and challenging task of underground garage parking.

%%%%%%%%%%%%%%%%%%%%%%%%%%%%%%%%%%%%%%%%%%%%%%
\section{Methodology}
\label{sec:methodology}
%%%%%%%%%%%%%%%%%%%%%%%%%%%%%%%%%%%%%%%%%%%%%%

\subsection{Planning System Design}
\label{Planning System Design}

\subsubsection{Overall Workflow}
\label{Overall Planning Process}

This paper proposes an automatic parking planning and control method based on an improved A* algorithm. The complete planning process is as follows:

\begin{enumerate}
    \item Load the grid map generated from real-world environmental information.
    \item Establish a vehicle kinematics model and use it to plan a parking path. 
    This step will be elaborated in \autoref{Vehicle Kinematic Model} and \autoref{Reverse Parking Planning Method}.
    \item Plan a path from the planning start point to the starting point of the parking path using the improved A* algorithm. 
    The principle of the A* algorithm will be introduced in \autoref{Principle of A* Algorithm}, while the specific optimization ideas of the algorithm involved in this paper will be detailed in \autoref{Algorithm Improvement Strategies}.
    \item After the path is generated, the MPC (Model Predictive Control) algorithm is used to achieve simulation control. 
    The MPC planning algorithm used in this paper will be explained in \autoref{MPC Control Algorithm}.
\end{enumerate}

\begin{figure}[htbp]
    \centering
    \includegraphics[width=1\linewidth]{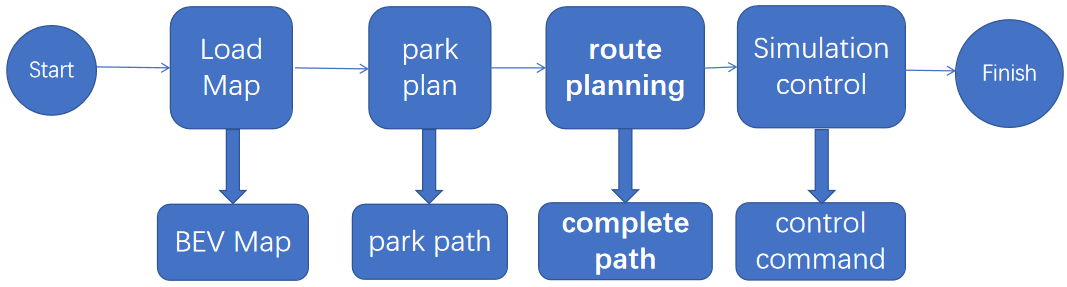}
    \caption{The pipeline of the proposed parking planning method. First, BEV map is loaded. Then generate the path with the improved A* algorithm. Finally, MPC algorithm generates control instructions}
    \label{fig:pipeline}
\end{figure}

The schematic diagram of the planning process is shown in \autoref{fig:pipeline}.

\subsubsection{Vehicle Kinematic Model}
\label{Vehicle Kinematic Model}

\paragraph{Kinematic Model}
\label{Kinematic Model}

To better simulate the vehicle's movement, this paper models the vehicle with the following assumptions:
\begin{enumerate}
    \item The vehicle does not have vertical movement.
    \item The two front wheels have the same or nearly the same angle and rotational speed, and the same applies to the two rear wheels.
    \item The steering angle of the front wheels controls the vehicle's yaw angle.
    \item The vehicle body and suspension are rigid bodies.
\end{enumerate}

Since each pair of front and rear wheels has the same state, it is possible to represent each pair of front and rear wheels as a single wheel based on the above assumptions. 
This paper uses four variables to describe the current state of the vehicle:

$x$: Horizontal ordinate of the vehicle's center
$y$: Vertical ordinate of the vehicle's center
$\psi$: Yaw angle of the vehicle, measured as the angle from the x-axis in a counterclockwise direction
$v$: Speed of the vehicle

\begin{figure}[!ht]  \centering
    \includegraphics[width=0.5\linewidth,trim={10 15 0 10},clip]{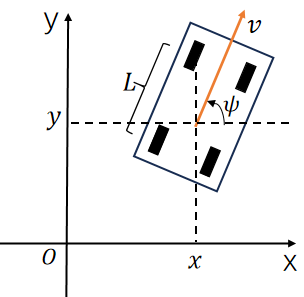}
    \caption{Vehicle Kinematic Model}
    \label{fig:Vehicle_Kinematic_Model}
\end{figure}

From this, the state vector of the vehicle can be derived:
\begin{equation}
z=\left[ x,y,v,\psi\right]
    \label{equ:state vector}
\end{equation}

When the vehicle is moving, let the steering angle of the front wheels be denoted as $\delta$. According to the model assumptions, the steering angle of the vehicle is also this value.
This study assumes that the rear wheel steering angle is 0, meaning the rear wheels always face the same direction as the vehicle body. 
The slip angle refers to the angle between the vehicle's velocity direction and the body orientation.
Let the slip angle be denoted as $\beta$.
In low-speed scenarios like parking, $\beta$ can be considered extremely small and neglected, so this paper assumes $\beta = 0$.
\begin{figure}[!ht] \centering
    \includegraphics[width=.95\linewidth,trim={150 140 150 145},clip]{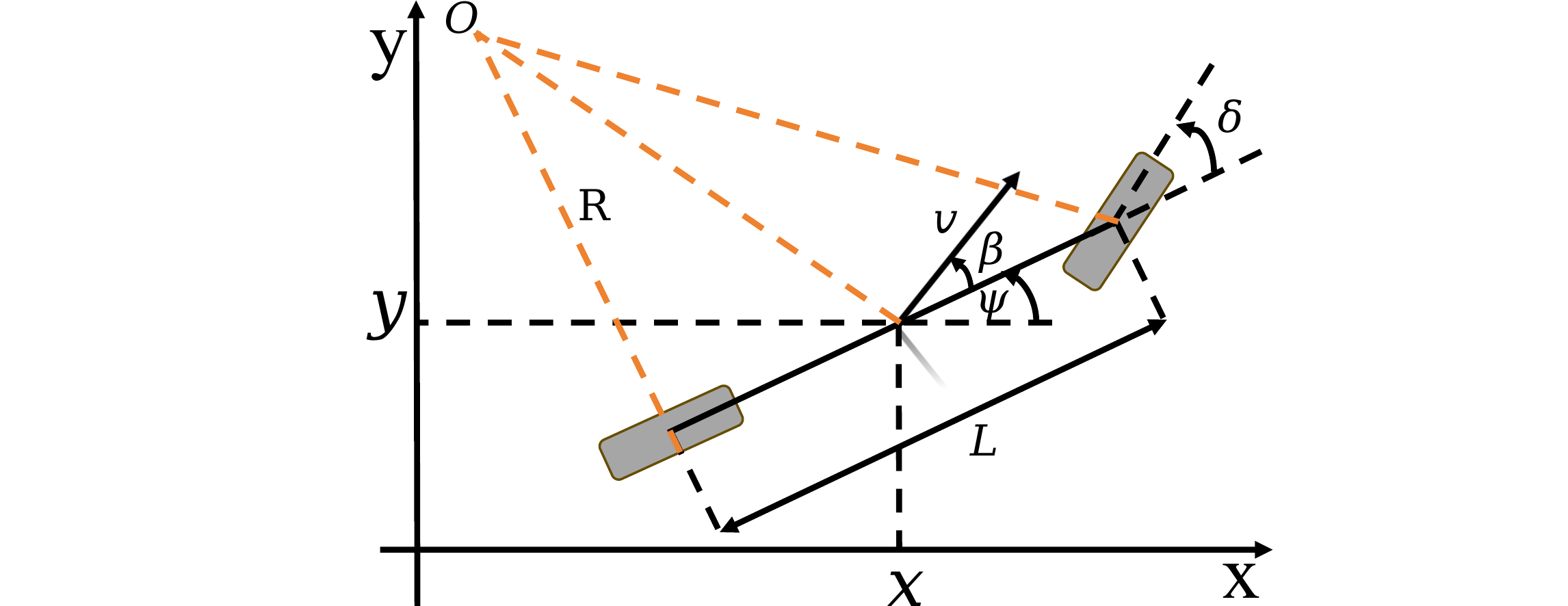}
    \caption{Kinematic Model of a Vehicle}
    \label{fig:Kinematic}
\end{figure}

The kinematic constraints for the front and rear wheels are as follows:
\begin{equation}
\begin{cases} \dot{x}\sin{(\psi+\delta)}-\dot{y}\cos{(\psi+\delta)}  = 0\\\dot{x}\sin\psi-\dot{y}\cos\psi = 0\end{cases} 
    \label{equ:Kinematics constraints of front and rear wheels}
\end{equation}

By solving these equations, we obtain:
\begin{equation}
    \begin{cases} \dot{x}=  v \cdot  \cos \psi\\\dot{y} = v \cdot  \sin \psi\end{cases} 
\end{equation}

Combining the above results, the model of the vehicle's state change at a given moment is:
\begin{equation}
\begin{cases} \dot{x}=  v \cdot  \cos (\psi)\\\dot{y} = v \cdot  \sin (\psi)\\\dot{v} = a\\\dot{\psi}= v \cdot  \tan (\delta)/L\end{cases} 
    \label{equ:Vehicle state change model}
\end{equation} 

Where $a$ represents the acceleration.

The planning and control algorithm presented in this paper uses this model to describe the vehicle's current state and its state changes during $\Delta t$.

\paragraph{Ackermann Steering Geometry}
\label{Ackermann Steering Model}

\begin{figure}[!ht]  \centering
    \includegraphics[width=.65\linewidth]{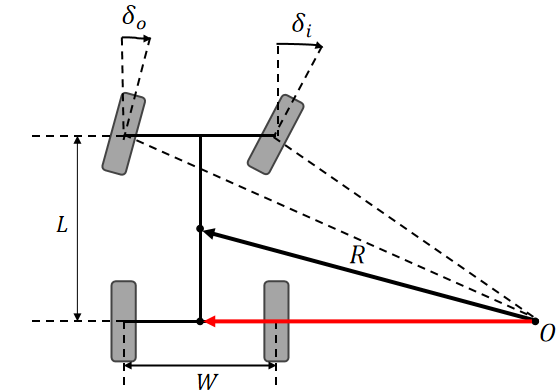}
    \caption{Ackerman steering model}
    \label{fig:Ackerman steering model}
\end{figure}

In \autoref{Kinematic Model}, it is assumed that the steering angles of the two front wheels of a vehicle are equal. 
However, in reality, these angles are not identical, and an average is taken to approximate them. 
Typically, the angle of the inner tire is slightly larger.
The situation of vehicle steering can be represented by the model shown in \autoref{fig:Ackerman steering model}. 
Here, the vehicle is making a normal right turn, with $L$ representing the wheelbase, $W$ representing the track width, $\delta_i$ representing the steering angle of the inner wheel, $\delta_o$ representing the steering angle of the outer wheel, $O$ representing the center of the vehicle's turning trajectory, and $R$ representing the radius of the vehicle's turning trajectory.
The rear wheels are assumed to maintain the same orientation as the vehicle.

After simplification, the average steering angle of the front wheels is denoted as $\delta= \frac{1}{2}(\delta_i+\delta_o)$, and the model is further simplified as shown in \autoref{fig:Ackerman steering model simplification}:

\begin{figure}[!ht] \centering
    \includegraphics[width=0.55\linewidth]{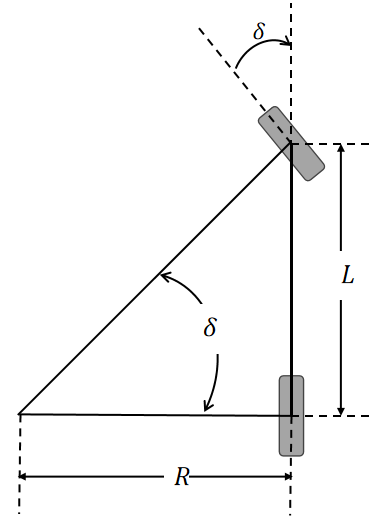}
    \caption{Ackerman steering model simplification}
    \label{fig:Ackerman steering model simplification}
\end{figure}

This allows us to obtain the minimum turning radius $R$ when the vehicle is steering.
\begin{equation}
R=\frac{L}{\tan{\delta}}
\end{equation} 

\subsubsection{Reverse Parking Planning Method}
\label{Reverse Parking Planning Method}

\paragraph{Parking Planning Process}
\label{Parking Planning Process}

Based on the vehicle dynamics model established in \autoref{Vehicle Kinematic Model} and the derived turning model, the minimum turning radius and the maximum turning capability of the vehicle can be calculated.
Consequently, this information can be utilized to plan a parking path that aligns with the vehicle dynamics, and the vehicle state can be described using a four-dimensional vector $z$.

The general procedure for the parking path planning method presented in this paper is as follows:

\begin{enumerate}
    \item Obtain the parking direction information and call different reverse parking planning methods according to whether the vehicle needs to be parked vertically or parallelly.
    \item Use geometric curves to plan the reverse parking path. Different paths are used for vertical and parallel parking.
    \item Set the starting point of the reverse parking path as the new endpoint for the complete path.
    \item Delegate the improved A* algorithm to complete the entire path planning task.
\end{enumerate}

It should be noted that using simple geometric curves to plan the parking path can result in curvature discontinuities at the junctions between straight lines and circular arcs, as well as at the connections of two circular arcs \cite{ref43}.
This situation does not meet the requirements for parking control. 
Therefore, it is necessary to use spline functions to fit and optimize the junctions of the path segments.

\paragraph{Vertical Parking}
\label{Vertical Parking}

For the convenience of evacuation, anti-theft, and easy charging for new energy vehicles, it is often preferred to park a car with the front facing outwards when parking in a perpendicular parking space.
Therefore, the vertical parking scenario proposed in this paper refers to: a car parking into a standard vertical parking space with its front facing outwards by reversing. 
This method is undoubtedly concise and effective, and its rationality has been verified in multiple papers [44].

The planned path is divided into two parts:

\begin{enumerate}
    \item Utilizing the car's dimensional information, the obstacle information near the parking space, and the turning radius calculated using the Ackermann steering geometry, an obstacle-free path is planned to allow the car to drive out of the parking space.
    \item A straight line is connected to the end of the path, and then smoothed using a Bezier curve, serving as a safeguard path.
\end{enumerate}

By concatenating and reversing the two path segments, the path for vertical parking is formed. This path ensures collision-free navigation and compliance with kinematic constraints. 
The endpoint P of the safeguard path becomes the new endpoint for the path planner.

The necessity of planning a safeguard path lies in providing the car with sufficient space to complete the reversing maneuver through continuous control commands.

\begin{figure}[!ht]   \centering
    \includegraphics[width=0.9\linewidth]{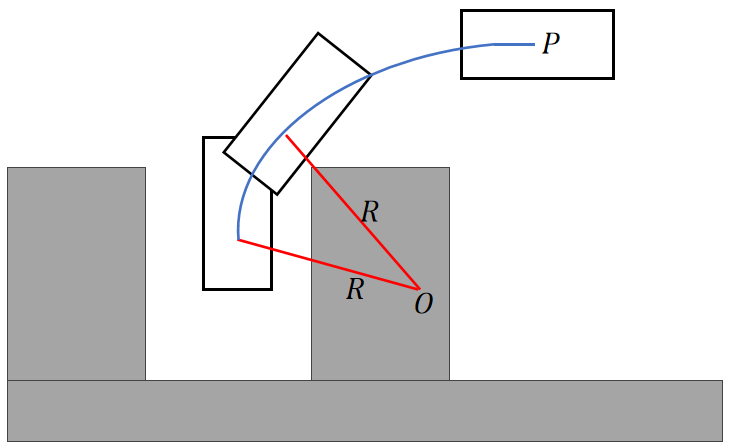}
    \caption{Vertical parking path}
    \label{fig:vertical_parking}
\end{figure}

\paragraph{Parallel Parking}
\label{Parallel Parking}

The same approach of retrieving the vehicle's exit path based on the parking space and reversing it is also adopted for parallel parking.
Additionally, parallel parking requires both continuous parking trajectories and no collisions with the parking space. The reversing trajectory can be simplified as \autoref{fig:parallel_parking}:

\begin{figure}[!ht] \centering
    \includegraphics[width=1\linewidth]{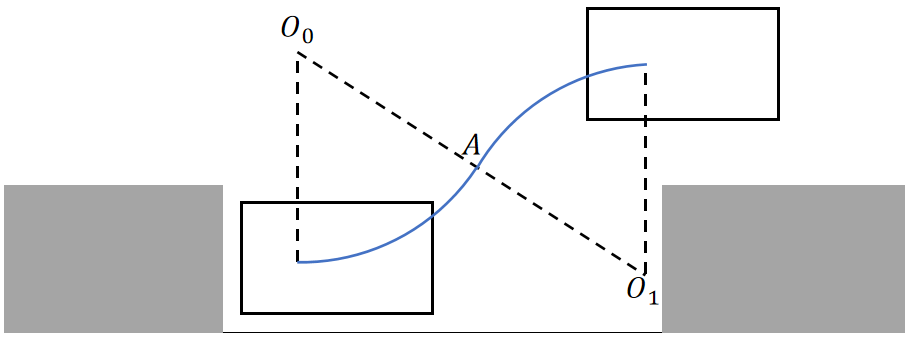}
    \caption{Parallel parking path}
    \label{fig:parallel_parking}
\end{figure}

Based on the above model, the vehicle's exit path during parking can be simplified into two tangent arcs. During specific planning, this method will add a straight line segment at the end of the exit trajectory, similar to \autoref{Vertical Parking}, and smoothen the entire path using Bezier curves.

For vertical parking, the legality of the parking space itself can be judged by calculating the length of the space. However, parallel parking requires additional calculation of the length constraint of the parking space. According to the Ackermann steering model mentioned in \autoref{Ackermann Steering Model}, the car needs to meet the minimum turning radius limitation when turning. At this time, the instantaneous rotation center lies on the same line as the rear wheels' connection \cite{ref45}. Meanwhile, the parallel parking scenario also needs to consider the maximum turning radius constraint. Taking the instantaneous rotation center at the start as the center, a circle with a certain radius is drawn. When the vertex $A$ of the car and the boundary $P$ of the parking space are on the same arc, the radius minus half of the car's wheelbase represents the maximum turning radius $R_{max}$. A sufficient condition for a legal parallel parking space is that the car's turning radius is less than the maximum turning radius. Therefore, if the minimum turning radius $R$ calculated based on the Ackermann steering model is still greater than $R_{max}$, it indicates that the car cannot park into the space without collisions.

\begin{figure}[!ht] \centering
    \includegraphics[width=0.9\linewidth,trim={50 30 30 3},clip]{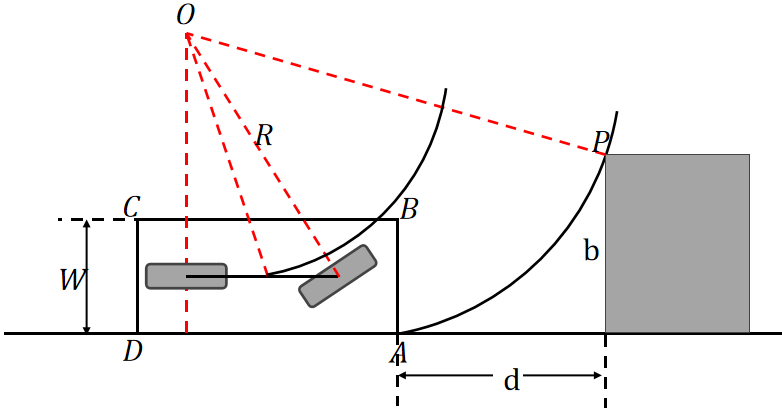}
    \caption{Collision constraint}
    \label{fig:Collision_constraint}
\end{figure}

Under the condition of satisfying the constraint, two arcs can be planned as the parking path. It is necessary to note that using methods such as Bezier curves to fit the collision-free parking path into the final path is essential. This is because there will be a sudden change in the steering angle at the joint of the two arcs.

\subsubsection{Principle of A* Algorithm}
\label{Principle of A* Algorithm}

\paragraph{Introduction to A* Algorithm}
\label{Introduction to A* Algorithm}

In \autoref{Reverse Parking Planning Method}, this method has planned the path for the vehicle to reverse and park into a parking space, and obtained the starting point of the parking path.
However, there is still a distance between the planning starting point and the starting point of the parking path. 
This solution adopts an improved A* algorithm, using the planning starting point as the origin and the starting point of the parking path as the destination, to complete the planning of this distance.
Then, the two path segments are concatenated and optimized comprehensively to obtain the final planned path.

The A* algorithm was first proposed in the last century \cite{ref46} and has been applied in various fields such as automatic navigation, pathfinding for space exploration rovers, electronic games, and more since its inception.
The A* algorithm combines a heuristic method similar to BFS with the conventional graph search algorithm, Dijkstra's algorithm.
The BFS heuristic method, which fully adopts a greedy strategy, often falls into local optimal solutions but has fast computational speed. 
Dijkstra's algorithm has slow computational speed when the problem size is large, but it can guarantee the optimal solution. 
The A* algorithm combines the advantages of both, ensuring that the shortest path is generated in less time than Dijkstra's algorithm.

\paragraph{Heuristic Evaluation Function}
\label{Heuristic Evaluation Function}

The A* algorithm is essentially an optimized Dijkstra graph search algorithm.
For search algorithms, to reduce the number of useless node expansions during the search for the best path, the algorithm must be able to intelligently select the "best" node. 
Therefore, an efficient search algorithm needs a way to evaluate the current node to be checked and then find a better node, rather than traversing all nodes to be checked aimlessly.

The A* algorithm achieves selective search by evaluating the node's quality through a heuristic evaluation function. This aspect makes its performance superior to the Dijkstra algorithm, with faster speed and fewer searched nodes.

The A* algorithm calculates node priority using the following function:
\begin{equation}
f(n)=g(n)+h(n)
    \label{equ:Heuristic_1}
\end{equation}
Where:
$f(n)$ represents the total cost of the node. 
When calculating the next search target, the A* algorithm always selects the node with the lowest total cost among the nodes to be checked.

$g(n)$ represents the cost from the starting point to node n, which is the cost already incurred from the planning start point to node n.

$h(n)$ represents the estimated cost from node n to the planning endpoint, serving as the heuristic function of the A* algorithm.

In grid maps, three heuristic functions are often used to represent the estimated cost, and different heuristic functions need to be selected based on the movement characteristics of the navigating object.

\begin{enumerate}
    \item 
Manhattan Distance

Manhattan distance represents the total length of the path from one point to another when only moving up, down, left, or right is allowed.

The Manhattan distance calculation formula is:
\begin{equation}
L=|x_1-x_2|+|y_1-y_2|
    \label{equ:Manhattan distance}
\end{equation}

Illustrate with a figure showing the path from red (start) to blue (end) using Manhattan distance:

\begin{figure}[htbp]
    \centering
    \includegraphics[width=0.35\linewidth]{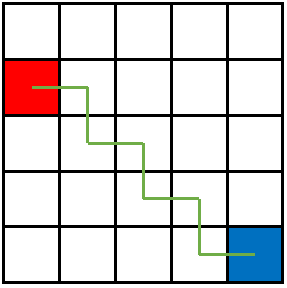}
    \caption{Manhattan Distance generated Path}
    \label{fig:Manhattan_distance}
\end{figure}

    \item 

Diagonal Distance

Diagonal distance represents the total length of the path from one point to another when moving in eight directions is allowed. 
The path using diagonal distance is illustrated below:

\begin{figure}[htbp]
    \centering
    \includegraphics[width=0.35\linewidth]{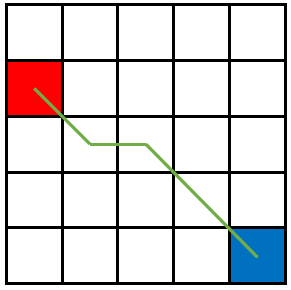}
    \caption{Diagonal Distance generated Path}
    \label{fig:Diagonal_distance}
\end{figure}

    \item 

Euclidean Distance

Euclidean distance represents the total length of the path from one point to another when moving in any direction is allowed.

The Euclidean distance calculation formula is:
\begin{equation}
L= \sqrt{(x_1-x_2)^2+(y_1-y_2)^2}
    \label{equ:Euclidean distance}
\end{equation}

Euclidean distance is the common distance calculation formula in daily life, and the illustration of the Euclidean distance path is as follows:

\begin{figure}[htbp]
    \centering
    \includegraphics[width=0.35\linewidth]{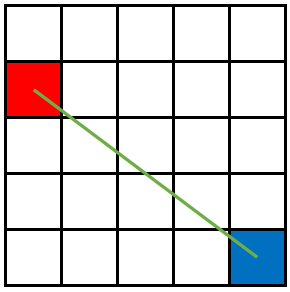}
    \caption{Euclidean Distance generated Path}
    \label{fig:Euclidean_distance}
\end{figure}

\end{enumerate}

Common implementations of the A* algorithm will select one of these three distances based on the specific situation of the planning problem, set up a heuristic function, and calculate the node cost accordingly.

\paragraph{A* Algorithm Process}
\label{A* Algorithm Process}

A* algorithm performs the search process and path generation for nodes in a map by maintaining two node lists - the open list and the closed list. 
The overall process of the algorithm is outlined as follows:

\begin{enumerate}
    \item Add the planning start point to the open list.
    \item Repeat the following steps until a stopping condition is met:

    \begin{enumerate}
        \item [a)] Iterate through the open list, find the node with the lowest $F$ value, and set it as the current node. 
        The $F$ value represents the total cost of the node, as mentioned in \autoref{Heuristic Evaluation Function}.
        \item [b)] Remove the currently searched node from the open list and add it to the closed list.
        \item [c)] Examine all adjacent reachable nodes of the current node, handling them differently based on the situation:

        \begin{enumerate}
            \item [i.] If an adjacent node is unreachable or already in the closed list, ignore it.
            \item [ii.] If an adjacent node is not in the open list, add it to the open list, set its parent node as the current node, and calculate the node cost.
            \item [iii.] If an adjacent node is already in the open list, check its current node cost to determine if the path through the current node is better. 
            If it is, update the parent node of the adjacent node to the current node and update the node cost accordingly.
        \end{enumerate}
        
        \item [d)] Stop when one of the following conditions is met:

        \begin{enumerate}
            \item [i.] The planning destination is added to the open list, indicating that a path to the goal has been found.
            \item [ii.] The destination is not reached, and the open list is empty, indicating that there are no more expandable nodes, and the planning fails.
        \end{enumerate}
        
    \end{enumerate}
    
    \item Starting from the destination point, backtrack along each node's parent node until returning to the start point, thus obtaining the complete path from the start to the destination.
\end{enumerate}

This algorithm was significantly advanced at the time because it inherited the advantages of both Dijkstra's algorithm and BFS algorithm. 
However, today, the original version of the A* algorithm cannot meet the planning requirements of new scenarios. 
To adapt traditional methods to the automatic parking planning task under local maps, this study decided to select the A* algorithm as the foundation and improve it, focusing on enhancing its planning speed and trajectory quality to meet the specific requirements of automatic parking tasks.

The improved A* algorithm proposed in this paper will optimize each step of the above process. 
For step 2a), this paper recalculates the node cost by optimizing the heuristic function introduced in \autoref{Heuristic Function Optimization} and reduces the time complexity of the algorithm using the binary heap optimization described in \autoref{Binary Heap Optimization}. 
For step 2c), the neighborhood expansion method is used to increase the search range of each search, reducing the number of searches and the total number of path nodes. 
This step will be introduced in \autoref{Neighborhood Expansion}. 
Additionally, the self-vehicle volume is introduced, and the node reachability is calculated by dynamically loading the obstacle map, allowing the planned path to consider the vehicle's size and improve planning speed. 
The optimization content will be specifically presented in \autoref{Introducing Vehicle Volume}. 
For the complete search process of the A* algorithm, the bidirectional search method introduced in \autoref{Bidirectional Search} is adopted to effectively reduce the problem scale. 
After path planning is completed, Bézier curve optimization described in \autoref{Bezier Curve Trajectory Optimization} is used to improve the trajectory quality of the complete planned path. 
Moreover, the method described in this paper also supports planning when the destination is illegal, which will be explained in \autoref{Unreachable Parking Spots}.

\subsubsection{MPC Control Algorithm}
\label{MPC Control Algorithm}

\paragraph{Control Signals}
\label{Control Signals}

After the planning algorithm generates an automatic parking path, this method utilizes the MPC control algorithm to achieve simulation control, track the vehicle's driving status, and evaluate the trajectory quality generated by the planning algorithm.

The kinematic model of the vehicle has been described in \autoref{Kinematic Model}. 
Based on this kinematic model, the control commands that the controller needs to output are:
\begin{equation}
u=\left[ a,\delta\right]
    \label{equ:Control signal}
\end{equation}

Where $a$ is the acceleration of the vehicle, and the controller adjusts the engine output and brake force to control the vehicle's acceleration. 
$\delta$ is the steering angle of the vehicle, and the controller adjusts the steering wheel and tire angles to control the vehicle's steering angle. 
Therefore, the controller can use these two variables to describe the vehicle's control input.

\paragraph{Introduction to MPC Control Algorithm}
\label{Introduction to MPC Control Algorithm}

This paper employs Model Predictive Control (MPC) as the autonomous driving control algorithm.

The MPC algorithm mainly consists of three steps:

\begin{enumerate}
    \item Based on the current state and system model, predict the system's state over a future period.
    \item Using the prediction results, numerically optimize the control sequence to find an optimal set of control sequences.
    \item Apply the first generated control command as the actual control signal.
\end{enumerate}

These three steps are repeated at each sampling time, performing optimization and prediction cyclically.
New measurement data will be used to update the optimization problem and solve it again, ensuring that the MPC algorithm can respond to system changes in real-time.
The constantly updated data and states guarantee the real-time and accuracy of the MPC algorithm.

The main advantage of MPC compared to traditional control methods lies in its high real-time performance. 
Instead of using a global optimization objective to calculate all control commands, predictive control methods calculate the current real-time optimal strategy separately for each time period. 
In contrast, traditional methods typically pre-solve a set of control commands and apply the entire set to system control. 
Even if external conditions change or the system fails to perfectly follow the control commands, traditional methods cannot make timely corrections. 
MPC is more flexible and adaptable, better able to cope with system uncertainties.

\begin{figure}[!ht] \centering
    \includegraphics[width=0.6\linewidth]{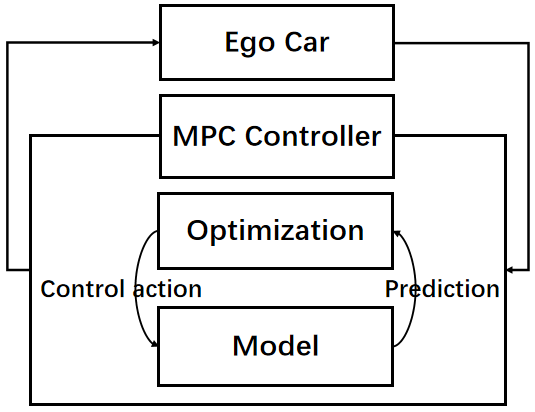}
    \caption{MPC flow chart}
    \label{fig:MPC}
\end{figure}

At each sampling time, MPC solves the optimization problem with the following objective function:
\begin{equation}
    \begin{array}{l}
    \min J(y, u)=\sum_{k=1}^{N}\left|y(t+k)-y_{d}(t+k)\right| \\
    \text { s.t. } \quad\left\{\begin{array}{l}
    y(t+1)=f(y(k), u(t)) \\
    u(t)=[a, \delta] \\
    a_{\min } \leq a \leq a_{\max } \\
    \delta_{\min } \leq \delta \leq \delta_{\max }
    \end{array}\right.
    \end{array}
\end{equation}

The meaning of the objective function is to minimize the difference between the system state and the desired state (i.e., the penalty function value) over the next time steps. 
This method introduces constraints on the vehicle control commands, considering the upper and lower limits. 
This also reflects real-world conditions—the vehicle cannot have arbitrarily large accelerations or steering wheel angles.

\paragraph{Control Requirements for Autonomous Parking}
\label{Control Requirements for Autonomous Parking}

Automated parking tasks are often completed in parking garages, such as underground garages.
\autoref{Characteristics of Autonomous Driving Parking Tasks} has already described the characteristics of typical scenarios for automated parking tasks, namely, narrow spaces, numerous obstacles, and directional restrictions for parking spaces. 
Therefore, the automated parking control system needs to optimize the output control instructions accordingly, ensuring that the vehicle's parking trajectory is continuous in curvature, smooth in operation, with minimal steering, and stable in driving trajectory.

To meet these control requirements, a penalty function for the MPC control algorithm is specifically designed in this paper. 
The aim is to enhance the performance of the control algorithm and achieve better parking control effects.

\paragraph{Cost Function Design}
\label{Cost Function Design}

Based on the control requirements of the automated parking task, the penalty function is designed with four components:

\begin{enumerate}
    \item Any acceleration or deceleration of the vehicle using the engine or brakes has a cost, i.e., the non-zero portion of $u$ has a cost.
    \item If the vehicle's motion state must be changed, it is desired that the difference between control instructions at consecutive sampling moments is minimized, i.e., $|u(t+1)-u(t)|$ has a cost.
    \item The actual position of the vehicle should be as close as possible to the position required by the planned path, thus $ \parallel D(t)-Z(t) \parallel_2$ should be included in the cost.  Here, $D(t)=\left[x_d(t) \quad y_d(t)\right]$ represents the expected position of the vehicle at time $t$ on the planned path, and $Z(t)=\left[x(t) \quad y(t)\right]$ represents the actual position of the vehicle at time $t$.
    \item When approaching the endpoint, it is desired that the vehicle aligns itself with the parking space, thus $|\psi-\psi_d|$ should be incorporated into the cost.  Here, $\psi_d$ represents the desired orientation of the vehicle when parking.
\end{enumerate}

From this, the objective function expression can be derived as:
\begin{equation} \small
\begin{split}
minJ(y,u)= \sum\limits_{k=1}^{N}W_1[u(t+k)]^2+ \\W_2\parallel [
u(t+k+1)-u(t+k)] \parallel_2 ^2+\\W_3\parallel [D(t+k)-Z(t+k)] \parallel_2^2+\\mW_4(\psi(t+k)-\psi_d)^2
\end{split}
    \label{equ:Objective function}
\end{equation}

Where, $W_1$,$W_2$ and $W_3$ are the weight matrices for the control cost, control instruction difference cost, and control position error cost, respectively.
$W_4$ is the weight for the pose error.$m$ determines whether to require the vehicle to quickly align its position, and is activated when the vehicle approaches the endpoint.
The expression is:
\begin{equation}
\begin{cases}m=0 & distance(now,goal) \geq d \\m=1 & distance(now,goal)  <  d\end{cases}
    \label{equ:m value}
\end{equation}

if the distance between the current vehicle position and the planned endpoint is less than $d$, $m$ take 1, urge the vehicle to adjust the position quickly.

During simulation experiments, the cost function is implemented using the Python language.

\subsection{Algorithm Improvement Strategies}
\label{Algorithm Improvement Strategies}

\subsubsection{Heuristic Function Optimization}
\label{Heuristic Function Optimization}

\paragraph{Search Tendency Optimization}
\label{Search Tendency Optimization}

The core of the A* algorithm that enables it to continuously expand nodes towards the goal and ultimately connect them into a complete path lies in the heuristic function that evaluates the future cost of a node. 
Therefore, if the optimal node always has a significantly better cost, it can reduce the algorithm's hesitation during the search process, enabling it to constantly progress along the optimal node without wasting time examining useless nodes.
Hence, the design of the heuristic function is the essential content of the A* algorithm.

In the physical world, a car has the ability to move in any direction, rather than being restricted to moving only up, down, left, or right. 
Therefore, in this study, among the three heuristic functions mentioned in \autoref{Heuristic Evaluation Function}, the Euclidean distance is chosen as the heuristic function to describe the estimated cost from the current node to the planning goal.

The method uses the calculation mentioned in \autoref{Heuristic Evaluation Function} to compute the node cost. 
Essentially, the node cost manifests as the 1:1 summation of the actual cost function $g(n)$ and the heuristic function $h(n)$.
If the weight ratio of these two components is adjusted, the behavior of the A* algorithm will also change accordingly.
Specifically, if the weight of the actual cost $g(n)$ is increased, the algorithm will tend to consider the actual cost of the path that has already been traversed, thus behaving more similarly to Dijkstra's algorithm. 
When the weight is significantly greater, the algorithm almost solely considers the actual cost while disregarding the estimated cost, and in this case, the A* algorithm can be considered to have degenerated into Dijkstra's algorithm. 
Conversely, if the weight of the heuristic function $h(n)$ is increased, the algorithm will tend to rely more on the estimated cost to guide the search direction.

Therefore, by adjusting the weight ratio between the actual cost and the estimated cost, this study can effectively control whether the A* algorithm's search behavior favors considering the actual cost or the estimated cost, thereby reducing unnecessary search points and improving search efficiency. 
Therefore, the heuristic function is expressed as:
\begin{equation}
    f(n)=g(n)+wh(n)
    \label{equ:Heuristic_2}
\end{equation}

Where $w$ is the weight coefficient, used to adjust the weight ratio of $g(n)$ and $h(n)$. This way, the algorithm can flexibly control the search tendency of the A* algorithm by adjusting the value of $w$.

At the beginning of the planning, when the current node is relatively far from the goal, it is unnecessary to try all possible paths to obtain the global optimum. 
Instead, the algorithm aims to quickly approach the goal with the shortest time consumption.
Therefore, at this stage, the algorithm tends to search using a greedy strategy to complete the planning task with minimal time consumption.
However, when approaching the goal, this method hopes that the algorithm can find the optimal path to meet the requirements of precise planning for automatic parking tasks and improve the trajectory quality.

The specific implementation strategy is that when the Euclidean distance from a new node to the planning goal is less than a certain threshold, $w<1$ is set to encourage the algorithm to generate an optimal path.
When the new node is far from the planning goal, $w>1$ is set to enable the path to quickly approach the goal. 
The specific weight values will be dynamically adjusted based on the experimental test results.

\paragraph{Node Comparison Optimization}
\label{Node Comparison Optimization}

\begin{figure}
    \centering
    \includegraphics[width=0.45\linewidth]{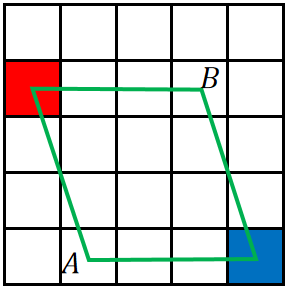}
    \caption{The algorithm gets tied up when multiple paths have the same cost}
    \label{fig:Repeat_search}
\end{figure}

One of the key factors that can degrade the performance of the A* algorithm is its tendency to get "tied up" when expanding nodes with the same $f(n)$ value. 
When multiple paths have the same current node cost, the algorithm will search these paths indiscriminately, even though in reality, only one of them needs to be explored. 
As shown in \autoref{fig:Repeat_search}, the algorithm may consider the paths from A and B to have the same cost, and thus attempt to plan to the goal from both of these paths sequentially.
To improve this situation, the algorithm needs to make a choice among nodes with the same heuristic function value to avoid unnecessary search overhead.

To achieve this selection, a slight offset $p$ can be added to the estimated cost $h(n)$. 
The role of this offset $p$ is to determine the priority by comparing the estimated costs $h(n)$ of nodes when their $f(n)$ values are the same.
Since nodes closer to the goal typically have smaller estimated function values $h(n)$, adding a slight offset $p$ to the estimated cost ensures that nodes closer to the goal have a slightly better heuristic function value $f(n)$, thus being prioritized by the algorithm. 
At the same time, because the offset is small, it does not significantly affect the established search tendency and interfere with normal node search.

Therefore, the cost function of a node can be expressed in form \eqref{equ:Heuristic_3}:
\begin{equation}
    f(n)=g(n)+(w+p) \cdot h(n)
    \label{equ:Heuristic_3}
\end{equation}

By doing so, the heuristic function can be further optimized, enabling the algorithm to quickly select one path among multiple nodes with the same cost, reducing unnecessary search behavior, and further improving the search efficiency of the A* algorithm.

\subsubsection{Binary Heap Optimization}
\label{Binary Heap Optimization}

In \autoref{Heuristic Function Optimization}, the design of the heuristic function is discussed.
In each step of the A* algorithm's expansion process, the cost of the nodes in the current open list is calculated based on the heuristic cost function, and the node with the lowest cost is selected for the next expansion.
Throughout the search process, the algorithm iterates over the entire open list multiple times, with each iteration having a time complexity of $O(n)$, resulting in significant overall time consumption.
Therefore, if the time complexity of selecting the optimal solution in each operation can be reduced, theoretically, the overall time efficiency of the algorithm can be improved by reducing the time spent searching for the optimal node in the open list.

A binary heap, as a special tree structure, is stored in an array form, with its root element always being the maximum or minimum value among all elements in the heap. The reason for choosing a binary heap lies in its ability to efficiently handle problems involving maximum or minimum values.

Both the insertion and deletion operations of a binary heap have a time complexity of $O(logn)$, and maintaining its relative order also requires only $O(logn)$ time.
By using a binary heap to store the open list, the operation of iterating through the entire table can be replaced by retrieving the root element and reorganizing the heap, reducing the time complexity of selecting the optimal node from $O(n)$ to $O(logn)$. 
This characteristic makes the binary heap play a significant role in reducing the time complexity of the A* algorithm.

The proposed method utilizes binary heap optimization to reduce the time complexity of selecting the optimal node in each step of the A* algorithm. 
The improved A* algorithm presented in this paper replaces the list with a min-heap data structure as the open list to store node data.
When selecting the optimal node, it only requires simply retrieving the root element of the min-heap and reorganizing the heap, without iterating through the entire open list. 
When adding a new node to the open list, there is also no need to sort the entire open list.

\subsubsection{Neighborhood Expansion}
\label{Neighborhood Expansion}

The improved A* algorithm selects the node at the top of the open list heap as the new current node.
Next, the adjacent nodes of the current node are checked to determine if they can be added to the open list. 
This paper adopts a neighborhood expansion method to adjust the adjacent node search strategy.

The idea of neighborhood expansion is to increase the search range during each node expansion to reduce the number of searches. 
The goal is to reduce the number of nodes on the final generated planning path and decrease the turning amplitude between nodes, thus making the planning path smoother.

Conventional A* algorithms often utilize four-neighborhood or eight-neighborhood searches.
When only allowing movement in the forward, backward, left, and right directions, Manhattan distance is used as the heuristic function, and the algorithm typically conducts a four-neighborhood search, attempting to add the four adjacent nodes above, below, left, and right to the open list in each search.
When allowing movement in eight directions, diagonal distance is used as the heuristic function, and the algorithm typically conducts an eight-neighborhood search, attempting to add the eight surrounding adjacent nodes to the open list in each search.

\begin{figure}[!ht]  \centering
    \includegraphics[width=0.45\linewidth,trim={0 0 0 0},clip]{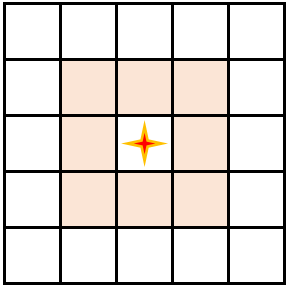}
    \caption{Traditional Eight-Neighborhood Search}
    \label{fig:8_nei}
\end{figure}

The schematic diagram of the traditional eight-neighborhood search is shown in \autoref{fig:8_nei}. 
Among them, the red star represents the location of the ego-vehicle on the grid map, and the pink squares represent the possible next positions to be searched.

\begin{figure}[!ht]    \centering
    \includegraphics[width=0.45\linewidth]{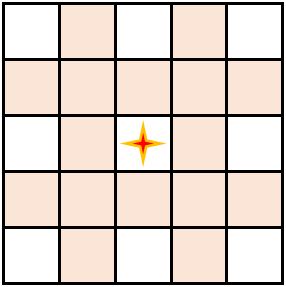}
    \caption{Nodes to be Searched after Neighborhood Expansion}
    \label{fig:16_nei}
\end{figure}

This algorithm performs a neighborhood expansion operation based on the traditional A* algorithm, expanding it to a sixteen-neighborhood search. 
The specific expansion method is shown in \autoref{fig:16_nei}. 
Since the automatic parking planning algorithm based on real-world scenarios does not need to be limited by the direction of movement. 
Real cars can make smaller turns when moving.

Although the neighborhood expansion method reduces the number of nodes in the final generated path, making it smoother, it increases the planning time consumption due to the need to search more adjacent nodes each time. 
Moreover, the neighborhood expansion planning method may lead the planner to generate phantom paths, requiring the car to pass through small obstacles. 
Therefore, whether to use neighborhood expansion should be considered based on specific usage scenarios. 
In the task studied in this paper, the map space is narrow with many obstacles, and high requirements for planning real-time performance.
Therefore, neighborhood expansion can be used as an optional strategy rather than the default strategy.

\subsubsection{Introducing Vehicle Volume}
\label{Introducing Vehicle Volume}

In \autoref{Neighborhood Expansion}, this paper discusses the selection strategy for the list of adjacent nodes. 
After obtaining the list of adjacent nodes for the current node, the A* algorithm determines whether these adjacent nodes are legitimate and can be added to the open list for further search.

There are two scenarios where an adjacent node is considered illegitimate:

\begin{enumerate}
    \item The adjacent node is already in the closed list. This indicates that the node has been searched before, and there is no need to redo the search.
    \item There is an obstacle at or near the adjacent node, preventing the ego-vehicle from reaching it.
\end{enumerate}

Traditional A* algorithms abstract robots as points without volume. 
However, in real automatic parking scenarios, cars definitely have a volume. 
Some situations that may occur in traditional A* algorithms, such as paths crossing corners or paths sticking closely to walls, are unacceptable in automatic parking tasks. 
If a car follows such a path, collisions or scratches are inevitable. Therefore, this method takes the car's volume into consideration.

\begin{figure}
    \centering
    \includegraphics[width=0.45\linewidth]{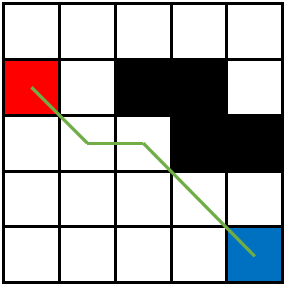}
    \caption{Trajectories closely sticking to walls and crossing corners}
    \label{fig:too_near}
\end{figure}

Consequently, the second scenario of illegitimate adjacent nodes in the improved A* algorithm is expanded to include: there are obstacles at or near the adjacent node, preventing the ego-vehicle from approaching it.

To determine whether a car can reach a node without an obstacle, the algorithm needs to further calculate the drivable area based on the obstacle map. 
As cars cannot actually reach areas too close to obstacles, the car's outline would collide or scratch if its center reached those areas.

\begin{figure}
    \centering
    \includegraphics[width=1\linewidth]{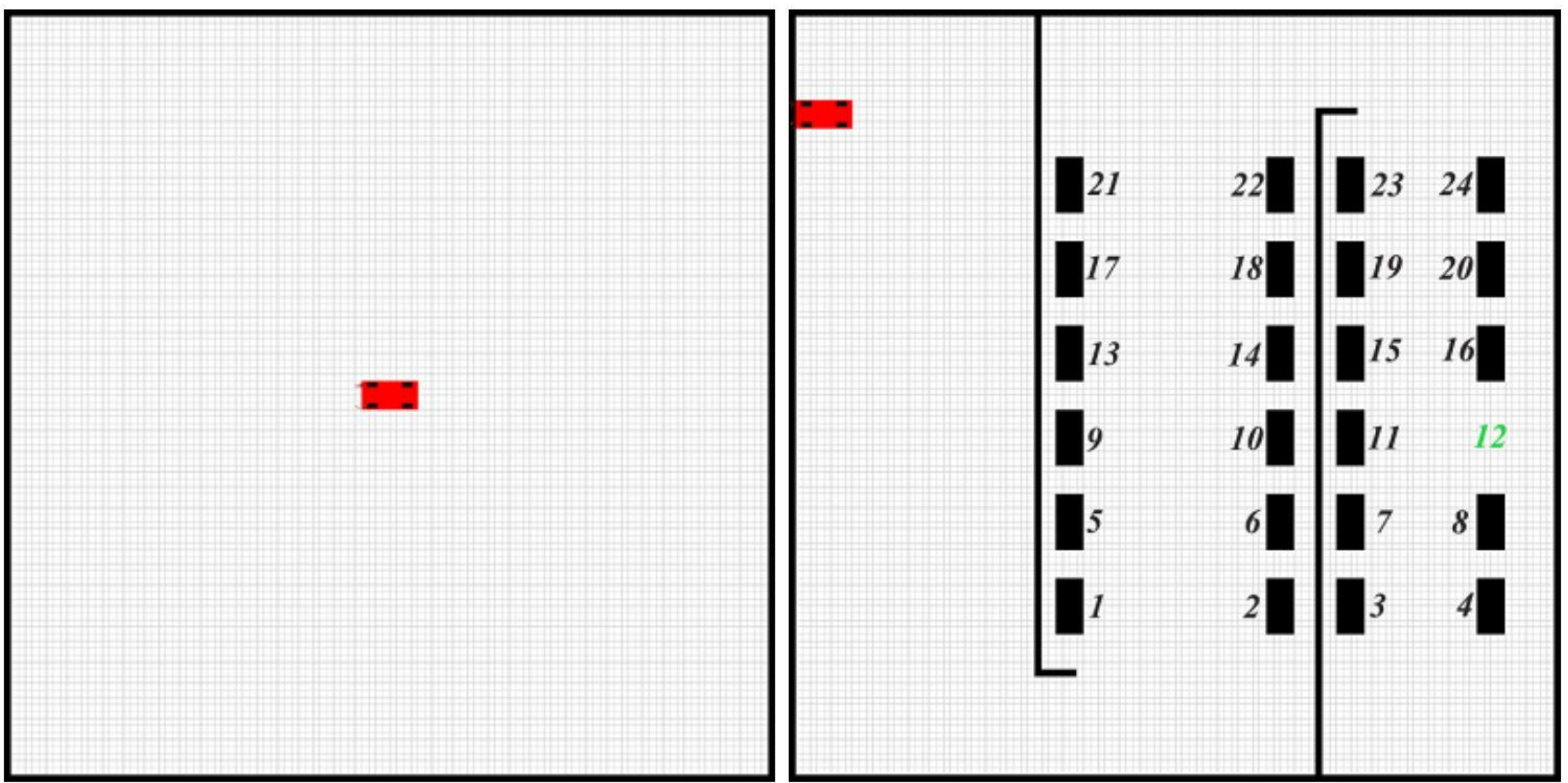}
    \caption{Sparse Obstacle Map}
    \label{fig:simple_map}
\end{figure}

In the 2020-2021 Rahneshan Autonomous Vehicle Competition, the winning team \cite{ref47} also took the ego-vehicle's volume into account during planning. 
Their method involved iterating through each position on the grid map during map loading, comparing it with stored obstacle information, drawing a circle with each grid point as the center and the car's size as the radius, and determining if there are obstacles near each position to calculate whether each position in the map allows the car to pass. 
This algorithm has a time complexity of $O(n^3)$, theoretically resulting in poor performance. 
However, the reason this method could be used is that the project employed a sparse obstacle map as shown in \autoref{fig:simple_map}, reducing the data scale to maintain the running speed within an acceptable range.

\begin{figure}[!ht] \centering
    \includegraphics[width=0.5\linewidth,trim={150 150 150 150},clip]{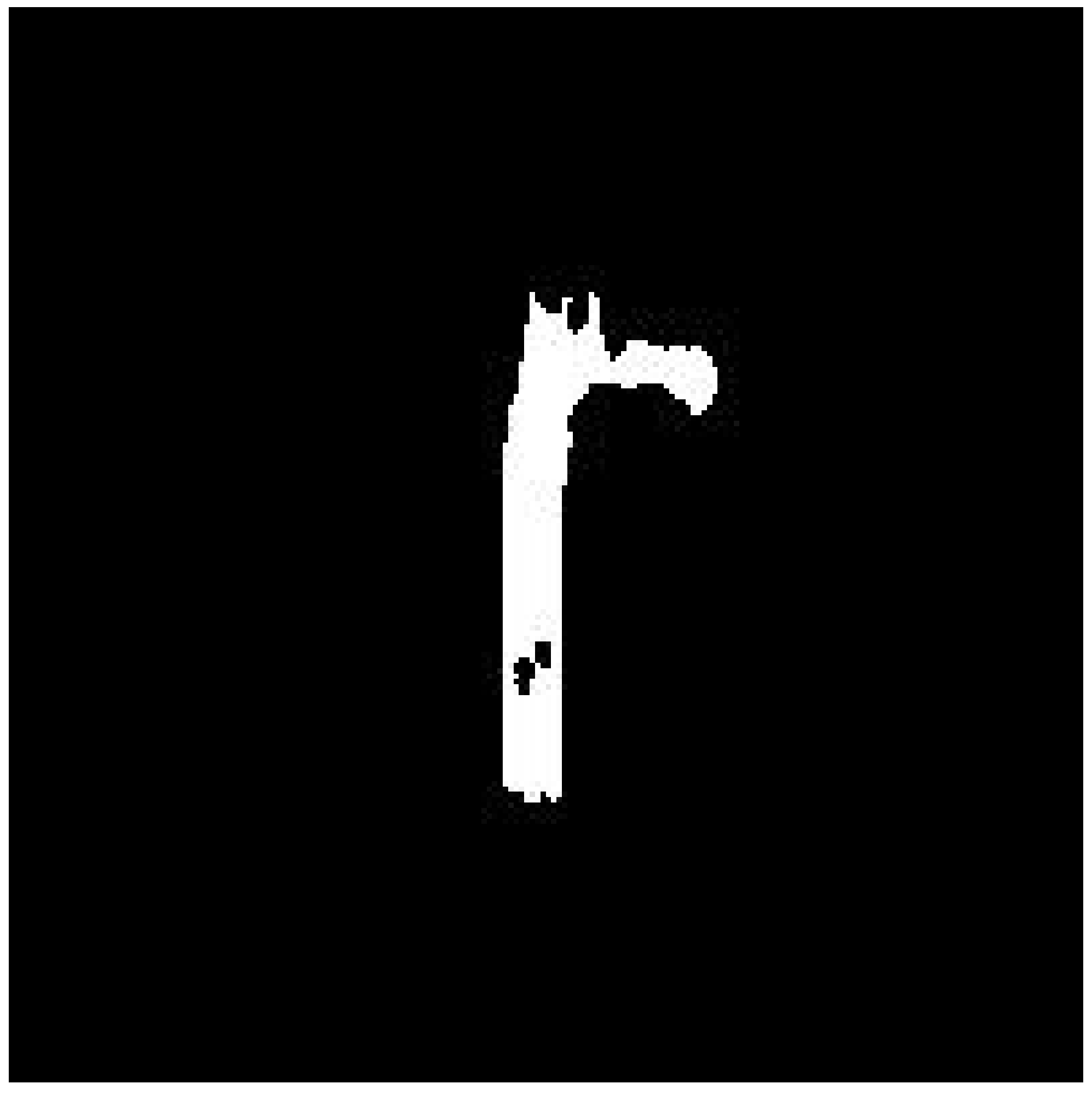}
    \caption{Grid Map Converted from a Real-World Scenario}
    \label{fig:grid_map}
\end{figure}

However, maps generated based on the real world contain far more obstacles than manually created simulation maps. 
Therefore, the time consumption of the algorithm used in this project would rapidly expand to an unacceptable level.
\autoref{fig:grid_map} shows a grid map converted from a real-world scenario, which is a local map generated by the ego-vehicle's perception module. 
The black areas represent non-drivable regions, while the white areas represent drivable roads. 
It can be observed that in automatic parking scenarios, the drivable areas are very narrow, and most of the area is occupied by various obstacles. 
Therefore, a more advanced algorithm must be used to calculate the traversability of nodes while considering the ego-vehicle's volume.

Cars are longer from front to back and narrower from left to right, but the need to open car doors must also be considered. 
Therefore, a circle is used as an approximation.
A necessary and sufficient condition for a node to be impassable is that there are obstacles within the range of a circle centered on that node and with a radius representing the size of the vehicle.

The proposed algorithm flow in this paper is outlined as follows:
\begin{enumerate}
    \item Initialize an obstacle matrix to indicate whether each grid cell on the grid map is traversable. 
    Each element in the matrix can have three possible values: -1, 0, and 1, representing impassable, unknown, and traversable, respectively.
    \item When expanding the current node, if the grid where the current node is located is recorded as traversable in the obstacle matrix, calculate the node cost normally; if it is recorded as impassable, ignore this node.
    \item If it is unknown whether the current node is traversable, take the current node as the center and the vehicle size as the radius, iterate through each point within the circle, and check whether these points have obstacles based on prior obstacle position information.
    If there is an obstacle at any point within the circle, record this grid as impassable in the obstacle matrix and ignore this node; if there are no obstacles within the circular area, record this grid as traversable in the obstacle matrix and calculate the node cost normally.
\end{enumerate}

The advantage of this algorithm is apparent: by integrating the calculation of node traversability with the A* algorithm's search process, it successfully achieves dynamic computation of the obstacle matrix.
This avoids calculating the traversability of all nodes, significantly reducing the problem scale and lowering the algorithm's time complexity.

\subsubsection{Bidirectional Search}
\label{Bidirectional Search}

The previous sections have implemented improvements to the search efficiency of the A* algorithm, but the overall problem size of the A* algorithm has not been reduced. 
Therefore, the A* algorithm is further optimized using a bidirectional search strategy.

Since the time consumption of the A* algorithm increases exponentially with the expansion of the problem size, an optimization method that can effectively reduce the search space and improve algorithm performance in theory is the bidirectional search algorithm. 
Bidirectional A* starts from the start and end points simultaneously, significantly reducing the problem size that each needs to handle.
The general algorithm flow is as follows:

\begin{enumerate}
    \item During initialization, the current node 1 starts searching from the start point towards the end point, and the current node 2 starts searching from the end point towards the start point.
    \item During each search process, the current node expands towards the node with the lowest cost in the closed list of the other side.
    \item If the two closed lists overlap during the search, recursively search for parent nodes from the overlapping point towards the planning start and end points to calculate the final path.
    \item If the open list of a current node is empty during the search, indicating that it cannot be further expanded, this means that no path can be found between the planning start and end points, and the algorithm terminates.
\end{enumerate}

The bidirectional A* algorithm and the basic A* algorithm have a significant difference besides the different problem sizes caused by unidirectional and bidirectional searches: the target point for expanding nodes in each step of the bidirectional A* algorithm is constantly changing.
When expanding nodes in the basic A* algorithm, the heuristic function always calculates the estimated cost from that node to the planning end point. 
However, in the bidirectional A* algorithm, each direction uses the estimated distance from the current node to the node with the lowest cost in the other direction as the heuristic cost. 
This difference results in a strong tendency for the two current nodes of the bidirectional A* algorithm to move closer to each other, ensuring that the two paths from the start to the end and from the end to the start of the bidirectional A* algorithm can definitely be connected into a complete path.

\subsubsection{Bezier Curve Trajectory Optimization}
\label{Bezier Curve Trajectory Optimization}

\paragraph{Bezier Curve}
\label{Bezier Curve}

This paper has discussed the optimization content of the improved A* algorithm applied in the planning process in \autoref{Heuristic Function Optimization} to \autoref{Bidirectional Search}. 
These optimizations reduce the algorithm's time complexity while also considering the actual size of the vehicle. 
However, the algorithm still relies on graph search on a grid map, resulting in paths filled with sharp corners, and discontinuities exist at the junctions of the planned path and the parking path.
To address these issues, the improved A* algorithm applies the extended Bézier curve method, specifically the B-spline curve, to interpolate and optimize the path after path planning.

Paths directly obtained from A* on a grid map often exhibit numerous sharp line segments and large turning angles. 
However, modern vehicles are unable to make such sharp turns, and these sharp lines will inevitably have a significant impact on the parking behavior.
To enable the vehicle to travel smoothly along the trajectory, it is necessary to smooth the path generated by the A* algorithm.
The purpose of curve smoothing is to eliminate unnecessary turns, reduce turning angles, optimize trajectory quality, and ensure continuous curvature, thus improving the usability of the path.

The Bézier curve is a tool for fitting line segments and can generate a fitting curve based on control points on the line segments.

Assuming that there are three points $P_0$, $P_1$, and $P_2$ forming a line segment, we wish to generate a curve based on these three points. In this case, $P_0$, $P_1$, and $P_2$ are the control points of the Bézier curve, and the generation of the curve is entirely related to these three points.

\begin{figure}
    \centering
    \includegraphics[width=0.55\linewidth]{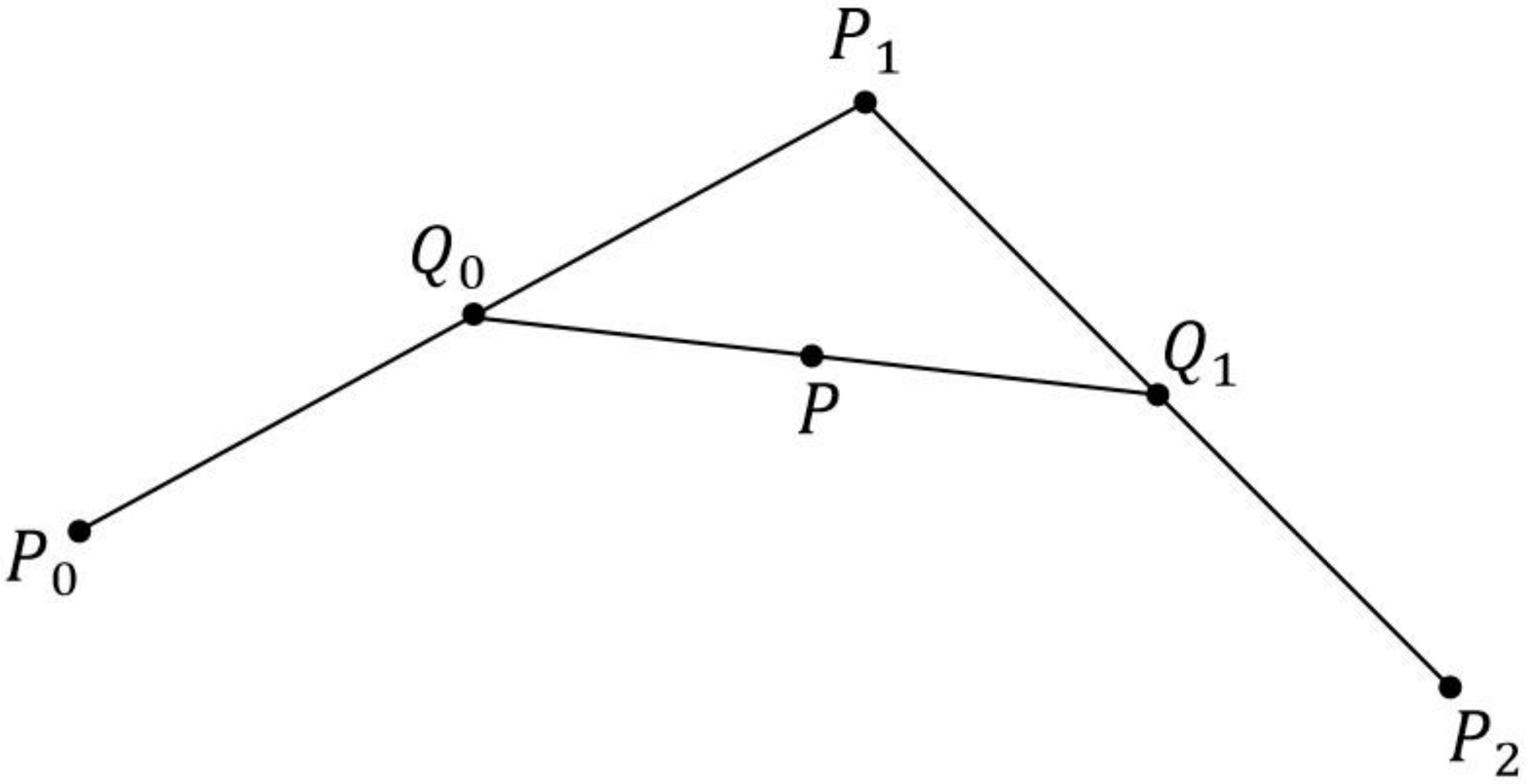}
    \caption{ELine segment formed by three points}
    \label{fig:Bezier_control_point}
\end{figure}

Taking a point $Q_0$ on $P_{0}P_{1}$, a point $Q_1$ on $P_{1}P_{2}$, and connecting them, then taking a point on $Q_{0}Q_{1}$ such that:
\begin{equation}
    \frac{P_0Q_0}{P_0P_1}=\frac{P_1Q_1}{P_1P_2}=\frac{Q_0P}{Q_0Q_1}
\end{equation}

We successfully use the positional information of $P_0$, $P_1$, and $P_2$ to describe the positional information of point $P$. 
The value of this equation represents the proportional distance of the moving point from a control point on the adjacent line segment.

By traversing all possible positions of the moving point, i.e., letting the value of the above equation take all numbers between 0 and 1, we have the following formula:
\begin{equation}
    \frac{P_0Q_0}{P_0P_1}=\frac{P_1Q_1}{P_1P_2}=\frac{Q_0P}{Q_0Q_1}=t,t\in\left[0,1\right]
\end{equation}

Connecting all positions of point $P$ at time $t\in\left[0,1\right]$ forms a curve, which is the Bézier curve.

\begin{figure}
    \centering
    \includegraphics[width=0.55\linewidth]{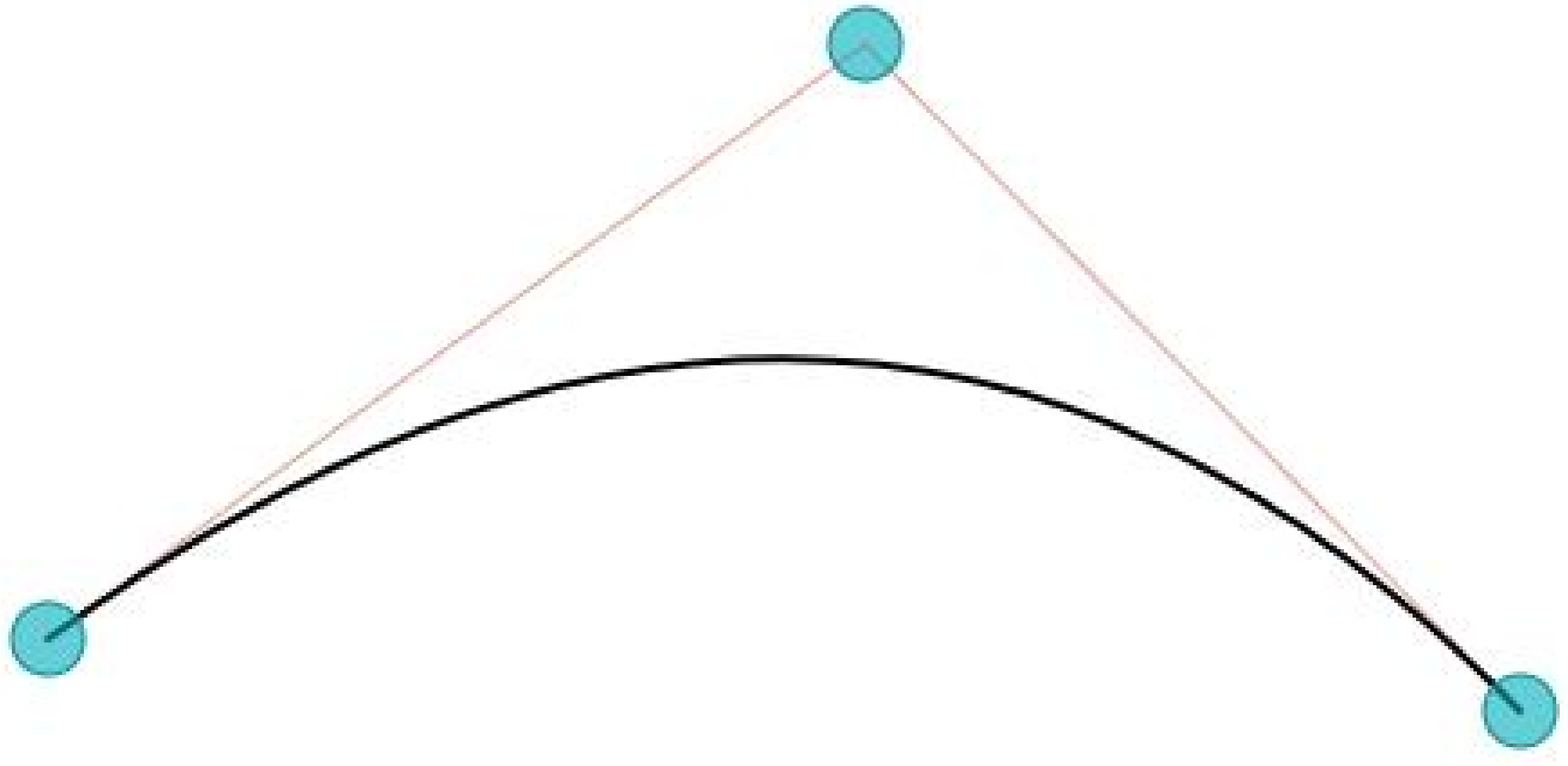}
    \caption{Drawing a Bézier curve with three control points}
    \label{fig:Bezier}
\end{figure}

When describing the position of each moving point individually, we have the following formula:
\begin{equation}
    \begin{split}
        Q_0=(1-t)P_0+tP_1 \\
        Q_1=(1-t)P_1+tP_2 \\
        P=(1-t)Q_0+tQ_1
    \end{split}
        \label{equ:Bézier_1}
\end{equation}

Extending this to cases with more control points, we can obtain the general formula for the Bézier curve:
\begin{equation}
    P(t)= \sum\limits_{i=0}^nP_iB_{i,n}(t),t\in\left[0,1\right]
    \label{equ:bezier curve}
\end{equation}

Where $B_{i,n}(t)$ is called the Bernstein basis function, and its expression is as follows:
\begin{equation}
B_{i,n}(t)= \left(\begin{array}{c}n\\ i\end{array}\right) t^i(1-t)^{n-i},i\in\left[0,\cdots,n\right]
    \label{equ:Bornstein basis function}
\end{equation}

\paragraph{B-spline Curve}
\label{B-spline Curve}

While the Bézier method indeed possesses numerous advantages, it also exhibits some limitations:

\begin{enumerate}
    \item The order of the Bézier curve is directly limited by the number of vertices in the characteristic polygon. 
    Once the number of vertices increases, the computational cost escalates rapidly.
    \item The Bézier curve demands high smoothness, which complicates the process of splicing. 
    Merging two Bézier curves can be challenging to achieve a smoothly connected new curve.
    \item Local modifications are not feasible. 
    Adjusting any part of the curve may affect the entire curve, resulting in a "butterfly effect" that is undesirable in certain application scenarios.
\end{enumerate}

However, the B-spline method successfully overcomes these three major drawbacks while inheriting all the advantages of the Bézier method. 
B-spline curves are extensions of Bézier curves, offering greater efficiency and flexibility in handling local modifications, complex curves, and large datasets.

The general formula for B-spline curves is as follows:
\begin{equation} \small
    P(t)= \sum\limits_{i=0}^nP_iB_{i,k}(t)
    \label{equ:B-spline}
\end{equation}
where
\begin{equation}
    \begin{array}{c}
    k=0, \quad B_{i, 0}(t)=\left\{\begin{array}{ll}
    1, & t \in\left[t_{i}, t_{i}+1\right] \\
    0, & \text { Otherwise }
    \end{array}\right. \\
    k>0, \quad B_{i, k}(t)=\frac{t-t_{i}}{t_{i+k}-t_{i}} B_{i, k-1}(t)+\\ \frac{t_{i+k+1}-t}{t_{i+k+1}-t_{i+1}} B_{i+1, k-1}(t)
    \end{array}
    \label{equ:B-base}
\end{equation}

The formula for B-spline curves is similar to that of Bézier curves, but it can be noticed that the subscript $n$ of the Bernstein basis function has been replaced by the $k$ of B-spline basis function, indicating that the degree of the polynomial of the B-spline is not related to the number of control vertices but is defined by the user. 
Moreover, the weights $t$ are no longer continuous values but a discrete list of control point positions. 
For instance, the control point list $t$ can take value as $\{0,\frac{1}{9},\frac{2}{9},\frac{3}{9},\frac{4}{9},\frac{5}{9},\frac{6}{9},\frac{7}{9},\frac{8}{9},1\}$. 
Consequently, $m+1$ nodes $t_0,t_1,\cdots,t_m$ successfully divide the curve into $m$ segments.

Utilizing B-spline curves, the path composed of multiple line segments generated by the A* algorithm can be effectively smoothed. 
The processed path ensures smoothness and continuity.

In this paper, a series of points are selected from the initially planned path with numerous unsmooth corners based on a predefined sampling rate, serving as control points for the B-spline curve.
Through these control points, the algorithm constructs a smooth B-spline curve. The interpolated and smoothed curve is then taken as the final path generated by the planning algorithm. 
This approach significantly improves the smoothness of the path generated by the A* algorithm, enhancing the driving efficiency and trajectory stability of the vehicle in automatic parking environments.

\subsubsection{Unreachable Parking Spots}
\label{Unreachable Parking Spots}

While the previous sections assumed that the input data for planning is legal and correct, this is not always the case in reality, and the proposed method needs to account for fault tolerance in case of illegal input data.

Due to the narrowness of underground garages, the automatic parking planning module often encounters situations where the parking spot is unreachable. 
This could be due to incorrect parking space coordinates output by upstream modules or too many obstacles near the parking space, leaving insufficient space for the vehicle to complete the reverse parking maneuver. 
Therefore, a strategy must be designed to enable the planning module to attempt parking planning even when the parking spot is unreachable.

This paper implements an attempt at planning when the parking spot is unreachable by maintaining a temporary optimal node during the A* algorithm search process. 
While expanding new nodes in the forward search, the algorithm determines whether the new node is closer to the planning goal than the previous nodes. 
If the newly expanded node is closer to the planning goal, it is updated as the temporary optimal node. 
If the planning algorithm fails to find a reachable path, it essentially degrades into a unidirectional A* search, continuously advancing until it finds an optimal temporary node. 
The algorithm then switches to using the temporary optimal node as the endpoint, gradually tracing back its parent nodes until it returns to the starting node.
This allows the algorithm to output a path that is as close to the planning goal as possible and navigate the vehicle to a position near the endpoint.

The implementation of this feature enables the planner to output a path even when the parking spot is illegal, rather than waiting indefinitely, enhancing the algorithm's ability to handle special situations.

\subsubsection{Summary of Optimizations for Improved A* Algorithm}
\label{Summary of Optimizations for Improved A* Algorithm}

This chapter summarizes the specific optimizations of the improved A* algorithm. The general process of the improved A* algorithm is as follows:

\begin{enumerate}
    \item Add the planning start point to the open list 1 for forward search expansion towards the planning goal; add the planning goal, i.e., the starting point of the parking path, to the open list 2 for backward search expansion towards the planning start point.
    \item For searches in each direction, repeat the following steps until a stopping condition is met:
    
    \begin{enumerate}
        \item [a)] Take the node at the top of the open list heap as the current node.
        \item [b)] Remove the current search node from the open list and add it to the closed list.
        \item [c)] Employ either the eight-neighbor or sixteen-neighbor expansion strategy to check all adjacent reachable points of the current node, handling different cases separately:

        \begin{enumerate}
            \item [i.] Detect the reachability of adjacent nodes based on the vehicle's dimensions. 
            Ignore the node if it is unreachable or already in the closed list.
            \item [ii.] If the adjacent node is not in the open list, add it to the open list, set its parent node as the current node, and calculate the node cost. 
            Set the heuristic function's target endpoint as the node with the lowest cost in the closed list of the other direction. 
            Adjust the heuristic function's weight value based on the distance from the target point.
            Reorganize the binary heap after adding the node.
            \item [iii.] If the adjacent node is already in the open list, check if the node cost needs to be updated. 
            Reorganize the binary heap after modifying the node cost.
        \end{enumerate}
        
        \item [d)] Stop when one of the following conditions is met:

        \begin{enumerate}
            \item [i.] The current nodes from both directions meet, indicating that a path to the target has been found.
            \item [ii.] The destination is not reached, and the open list is empty, indicating that the parking spot is unreachable. 
            In this case, continue the forward search, expanding nodes along the planning endpoint, and maintaining a temporary optimal node.
        \end{enumerate}
        
    \end{enumerate}
    
    \item Start the search in both directions from the current nodes, tracing back along each node's parent node until returning to their respective starting points. 
    Then, concatenate the two path segments generated by the bidirectional search to obtain the complete path from the planning start point to the starting point of the parking path. 
    If the parking spot is unreachable, backtrack from the temporary optimal node towards the planning start point.
    \item Concatenate the parking path. 
    \item Use the B-spline curve, a Bezier curve extension method, to interpolate and optimize the complete path.
\end{enumerate}

The improved A* algorithm proposed in this paper utilizes various optimization methods, effectively enhancing the quality of the generated trajectory and significantly reducing the computational time of the algorithm.

%%%%%%%%%%%%%%%%%%%%%%%%%%%%%%%%%%%%%%%%%%%%%%
\section{Experiments}
\label{sec:experiments}
%%%%%%%%%%%%%%%%%%%%%%%%%%%%%%%%%%%%%%%%%%%%%%

\subsection{Experimental setup}
\label{Experimental setup}

To validate the performance of the improved planning control algorithm based on the A* algorithm, this paper conducted multiple sets of simulation comparative tests. 
The purpose of these tests was to measure the improvement of the proposed algorithm compared to the traditional A* algorithm. 
The simulation experiments were implemented on an Ubuntu 20.04 LTS operating system based on the Linux kernel, using a computer with an 8-core CPU and a total of 32GB of available memory.
The algorithm was written in Python 3.8, using VS Code as the IDE, and miniconda was utilized to manage the code's working environment. 
All experiments in this paper were conducted under the same conditions to ensure fairness and comparability of the results.

For the experiments, a map generated from real-world scenarios was used as the grid map for the automatic parking planning algorithm. 
The map size was 200$\times$200. 
Before the experiments, the SurroundOcc \cite{ref48} artificial intelligence perception model developed by Tsinghua University and Tianjin University was used to perceive the pure visual scene data provided by the nuScenes dataset, generating BEV (Bird's Eye View) occupancy point clouds. 
These point clouds were then converted into grid maps for use in the simulation testing of the planning control algorithm.

\subsection{Experimental Process}
\label{Experimental Process}

The following procedure was adopted for the simulation experiments:
\begin{enumerate}
    \item Utilize the SurroundOcc model to generate real-world occupancy point clouds and convert them into grid maps.
    \item Select grid maps for vertical parking scenarios and parallel parking scenarios separately and use the selected grid maps as obstacle maps for the planning algorithm under test.
    \item Input the planning start coordinates, parking space center coordinates, and specify whether it is parallel or vertical parking.
    \item Execute the planning algorithm and output the planned path result graph.
    \item Implement the Model Predictive Control(MPC) algorithm to control the vehicle in the simulation environment, starting from the starting point and completing the parking path.
    \item Collect and evaluate the statistical parameters.
\end{enumerate}

\subsection{Evaluation Parameters}
\label{Evaluation Parameters}

The following parameters were used in this paper to evaluate the quality of planning:

\begin{enumerate}
    \item Planning time$(ms)$
    \item Total length of the generated path
    \item Average acceleration of the trajectory$(m/s^2)$
    \item Average steering angle of the trajectory$(\degree)$
\end{enumerate}

\subsection{Ablation Experiments}
\label{Ablation Experiments}

To validate the effectiveness of the multi-dimensional optimization method for the A* planning algorithm proposed in this paper, a series of ablation experiments were designed in this chapter. 
Ablation experiments are a method of evaluating the impact of certain components on the overall performance by gradually removing or adding them from the algorithm. 
The purpose of this experiment is to clarify the specific contributions of different optimization methods to the performance of the A* algorithm.

To clarify the contribution of each optimization method, the following ablation experiment scheme was designed:

\begin{itemize}
    \item Experiment 1: Without using heuristic weighting.
    \item Experiment 2: Without using dynamic obstacle loading.
    \item Experiment 3: Without using bidirectional A* and binary heap.
    \item Experiment 4: Using neighborhood expansion.
    \item Experiment 5: Without using Bézier curve optimization.
    \item Experiment 6: Without using Bézier curve optimization, bidirectional A* and binary heap optimization.
\end{itemize}

As a contrast, schemes with no optimization at all and schemes with comprehensive optimization are added to the statistical evaluation parameters of each method.

\begin{itemize}
    \item Baseline Experiment: Traditional A* algorithm with no optimizations.
    \item Improved A*: Results after comprehensive optimizations.
\end{itemize}

The maps, planning start points, planning end points, and parking methods used in the ablation experiments were all the same.

%%%%%%%%%%%%%%%%%%%%%%%%%%%%%%%%%%%%%%%%%%%%%%
\section{Results}
\label{sec:results}
%%%%%%%%%%%%%%%%%%%%%%%%%%%%%%%%%%%%%%%%%%%%%%

\subsection{Vertical Parking}
\label{Vertical Parking}

The following map depicts a vertical parking scenario, with the starting coordinate at (95,85) and the final parking spot coordinate at (109,133).

\begin{figure}[ht!]
  \centering
    \begin{subfigure}[t]{0.235\textwidth}
      \centering   
      \includegraphics[width=1\linewidth,trim={20 20 20 57},clip]{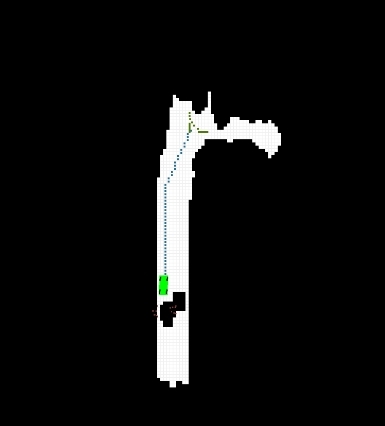}
        \caption{Basic A* algorithm planning result}
        \label{fig:A_V.sub1}
    \end{subfigure}
    \begin{subfigure}[t]{0.235\textwidth}
      \centering   
      \includegraphics[width=\linewidth,trim={20 20 20 55},clip]{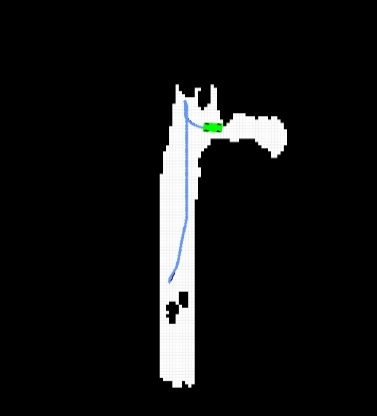}
        \caption{Improved A* algorithm planning result}
        \label{fig:A_V.sub2}
    \end{subfigure}
\caption{\label{fig:A_star_vertical}Comparison of Basic A* Algorithm and Improved A* Algorithm in Vertical Parking Scenario}
\end{figure}

The planning result using the basic A* algorithm is shown in \autoref{fig:A_V.sub1}.   
The original planning map size is 200×200 pixels, identical to the size of the SurroundOcc-generated perception point cloud.   
To make the planned path more visible, we have cropped the effective part of the planning result map.   
In the figure, the ego vehicle is represented in green, black indicates non-drivable areas (other vehicles, curbs, greenery, pedestrians, etc.), white represents drivable areas, cyan represents the path generated by the basic A* algorithm from the planning start point to the parking start point, and grass green depicts the path from the parking start point to the final parking spot.

It can be observed that there is an abrupt turn in the cyan path section, and a real car constrained by the minimum turning radius cannot make a perfect sharp turn.   
Additionally, the joint between the cyan path and the grass green parking path also has a relatively abrupt transition, and a discontinuous path would require the car to reorient its front wheels at each corner \cite{ref45}.   
This undoubtedly leads to a decline in performance.

% \begin{figure}[!ht] \centering  
%     \subfigure[Basic A* algorithm planning result]{
%         \label{fig:A_V.sub1}
%         \includegraphics[width=0.47\linewidth,trim={20 20 20 57},clip]{graphics/BasicA.jpg}}
%     \subfigure[Improved A* algorithm planning result]{
%         \label{fig:A_V.sub2}        \includegraphics[width=0.47\linewidth,trim={20 20 20 55},clip]{graphics/ImprovedA.jpg}}
%     \caption{Comparison of Basic A* Algorithm and Improved A* Algorithm in Vertical Parking Scenario}
%     \label{fig:A_star_vertical}
% \end{figure}

The planning result using the proposed improved A* algorithm is shown in \autoref{fig:A_V.sub2}.   
The optimized path after trajectory refinement is represented in blue, and in the simulation environment, the car smoothly follows the path to the destination.   
The optimized path no longer has harsh turns or discontinuous path joints with inconsistent curvature, and the trajectory quality is significantly improved.

According to the parameters proposed in \autoref{Evaluation Parameters}, the recorded data is as follows:

\begin{table*}[!ht]   \small   \centering
    \setlength\tabcolsep{2pt} \renewcommand{\arraystretch}{1.0}
    \caption{Algorithm Performance Comparison in Vertical Parking Scenario}
    \begin{tabular}{ccccccc}   \toprule
        algorithm & Map loading time & Planning time & Path length & Travel time & Average acceleration & Mean steering Angle \\ \midrule
        Basic A* & 98255.68 & 21.86 & 78.14 & 20.4 & 1.812 & 19.743 \\ 
        Improved A* & \textbf{2.062} & \textbf{411.04} & 72.119 & 49.6 & \textbf{0.306} & \textbf{10.791} \\ \bottomrule
    \end{tabular} 
    \label{tab:comparison_vertical}
\end{table*}

The improvement in the algorithm's performance is evident from the tabular data.
The optimized obstacle detection method and path planning approach have significantly enhanced the planning efficiency, resulting in a total time reduction from 98277.54 milliseconds to 413.072 milliseconds for loading the map and planning the path. 
In a real-world scenario of an underground parking garage, the planning speed has increased by over 95\%.
The basic A* algorithm was almost unable to plan an effective path within an acceptable timeframe under such a complex map, while the improved A* can provide a path in a shorter time, making it more practical.

Concurrently, the trajectory quality has also improved. 
Utilizing the MPC algorithm to control the vehicle in a simulation environment to complete the parking task, the vehicle traveled along different trajectories.
The average acceleration of the trajectory planned by the basic A* algorithm was 1.812, while the improved A* algorithm achieved an average acceleration of only 0.306, representing an optimization of 83.1\%. 
Similarly, the average steering angle of the trajectory planned by the basic A* algorithm was 19.743°, while the improved A* algorithm reduced it to only 10.791°, achieving an optimization of 45.3\%.

\begin{figure}[!ht]
    \centering
    \includegraphics[width=0.96\linewidth,trim={35 12 55 48},clip]{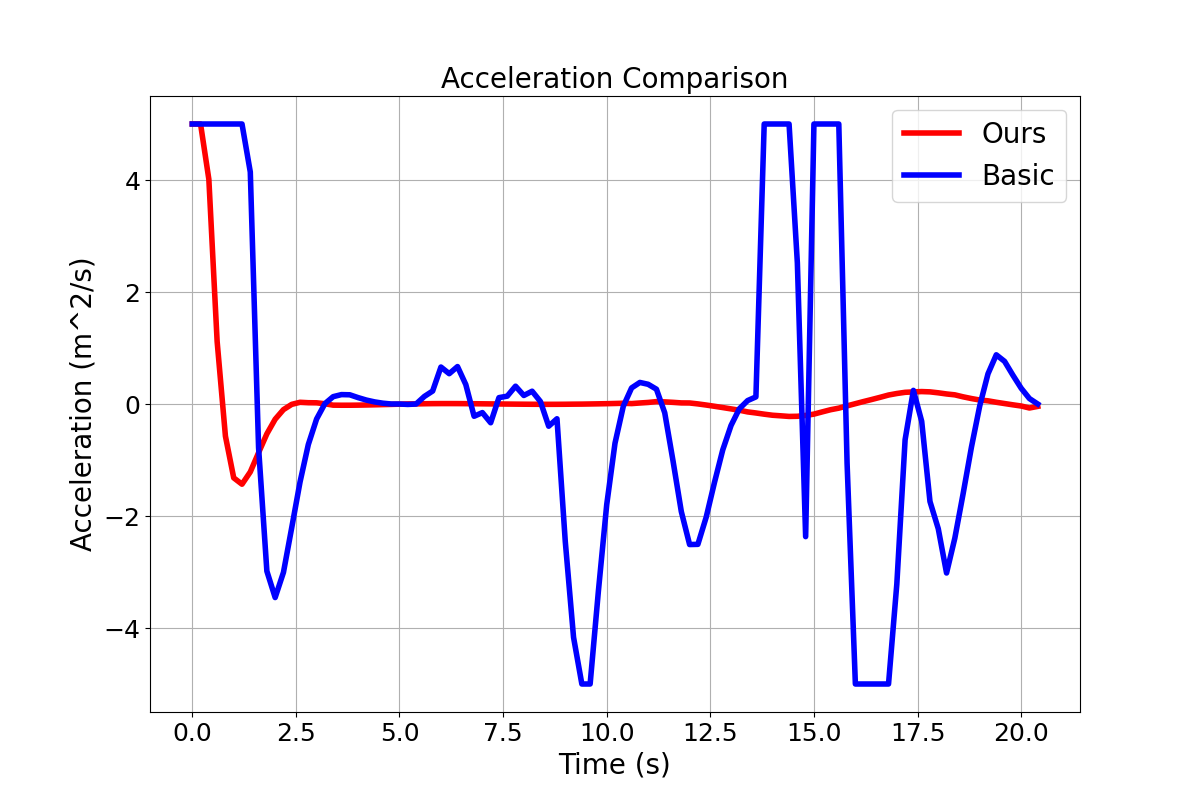}
    \caption{\label{fig:acc}Comparison of Accelerations for Navigation Paths Generated by Baseline A* and Improved A* Algorithms}
\end{figure}

The acceleration curves of the basic A* and improved A* algorithms during vehicle travel are shown in \autoref{fig:acc}. 
The horizontal axis represents the simulation time of vehicle travel, while the vertical axis represents the acceleration.
The sampling range covers the vehicle's travel behavior from the planning start point to the start of the parking completion path.
Among them, the blue curve represents the acceleration curve corresponding to the basic A* algorithm, while the red curve represents the acceleration curve corresponding to the improved A* algorithm proposed in this paper. 
It can be observed that the acceleration curve of the improved algorithm is very smooth, while the acceleration curve of the basic A* algorithm is highly unstable.
This is attributed to the fact that the planning path generated by the basic A* algorithm contains numerous sharp corners and discontinuous curvature points, forcing the controller to repeatedly accelerate and decelerate to adapt to the path.
However, the improved A* algorithm generates a smoother and more continuous path, thus eliminating the need for the controller to constantly adjust the vehicle's speed.

\begin{figure}[h]
    \centering
    \includegraphics[width=0.96\linewidth,trim={35 12 55 48},clip]{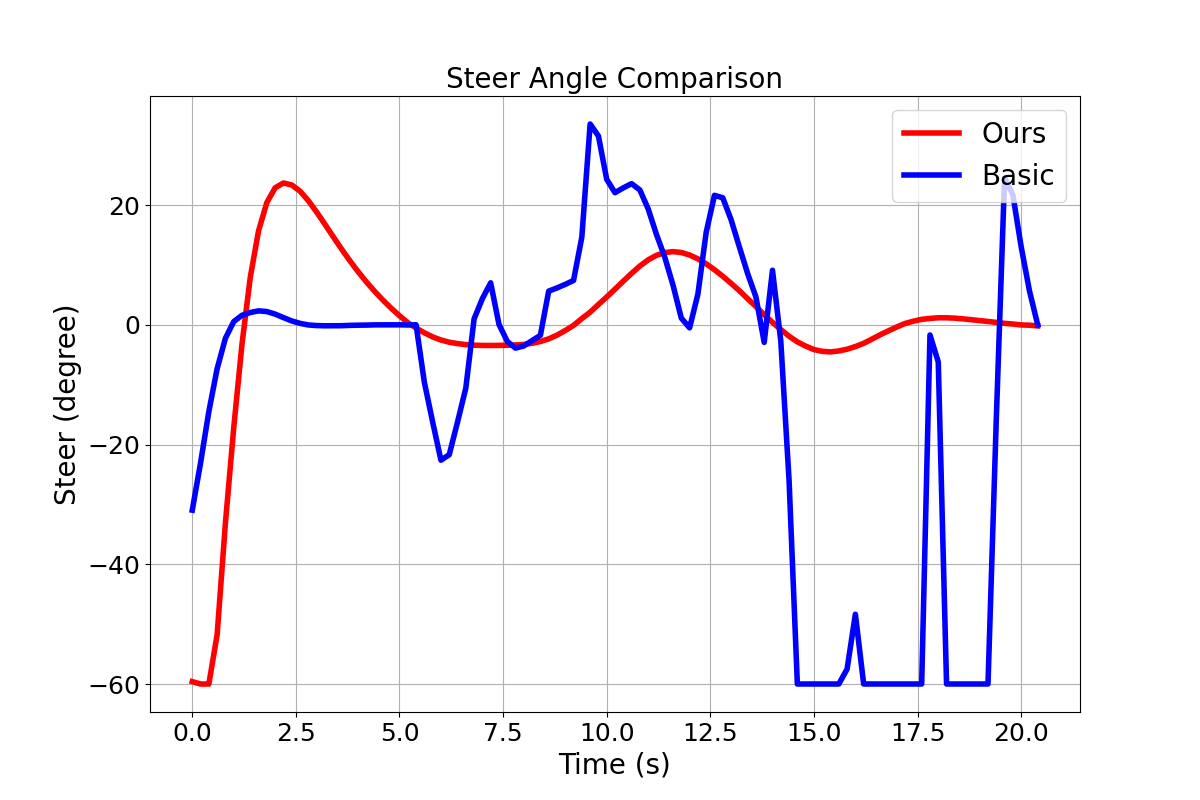}
    \caption{\label{fig:steer}Comparison of Steering Angles for Navigation Paths Generated by Baseline A* and Optimized A* Algorithms}
\end{figure}

\autoref{fig:steer} depicts the comparison of steering angles between the basic A* and improved A* algorithms. 
Similarly, it can be observed that the path generated by the basic A* algorithm prompts the planner to adopt violent driving behaviors, such as sharply turning the steering wheel.
In contrast, the path generated by the improved A* algorithm allows the planner to use smoother operations to control the vehicle's movement.

The statistical data and charts demonstrate that the improved A* algorithm generates planning trajectories with higher quality, shorter paths, and significantly reduced unnecessary acceleration/deceleration and steering wheel movements.
This results in a safer, more stable, and comfortable driving experience. 
Moreover, the improved A* algorithm significantly enhances efficiency, enabling it to adapt to complex underground parking scenarios and the high real-time requirements of local maps.

\subsection{Parallel Parking}
\label{Parallel Parking}

The following map depicts a parallel parking scenario, with the starting point at (110,65) and the target parking space located at (142,110).

\begin{figure}[ht!]
  \centering
    \begin{subfigure}[t]{0.235\textwidth}
      \centering   
      \includegraphics[width=1\linewidth,trim={120 0 30 120},clip]{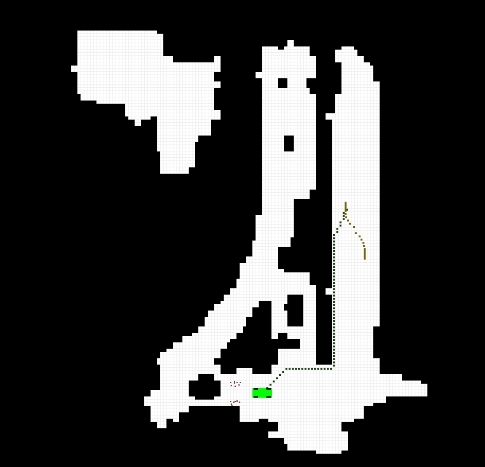}
        \caption{Basic A* algorithm planning result}
        \label{fig:A_P.sub1}
    \end{subfigure}
    \begin{subfigure}[t]{0.235\textwidth}
      \centering   
      \includegraphics[width=\linewidth,trim={120 0 30 120},clip]{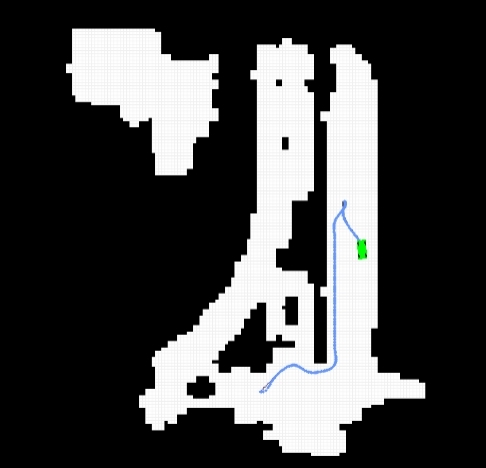}
        \caption{Improved A* algorithm planning result}
        \label{fig:A_P.sub2}
    \end{subfigure}
\caption{\label{fig:A_star_parallel}Comparison of Basic A* Algorithm and Improved A* Algorithm in Parallel Parking Scenario}
\end{figure}

The A* algorithm planning result is shown in \autoref{fig:A_P.sub1}. The trajectory generated by the basic A* algorithm is still represented in two randomly selected colors to emphasize the stitching traces of the path. 
It is worth noting that the traditional A* algorithm performs poorly in this scenario.
The planned path exhibits severely unrealistic driving behaviors. 
Excessive proximity to obstacles is a highly dangerous behavior that is unacceptable in real-world driving. 
Furthermore, the planned trajectory requires the vehicle to complete a right-angle turn and two sharp turns, with an additional sharp turn at the connection of the parking path, disregarding the physical constraints of the vehicle.

The planning result of the improved A* algorithm proposed in this paper is shown in \autoref{fig:A_P.sub2}. 
It can be observed that the optimized blue trajectory has smoother turns and also leaves space for the vehicle to stay away from the walls. 
The improved A* algorithm still outperforms the basic A* algorithm in this scenario, significantly reducing planning time while making the trajectory more continuous and smooth.

% \begin{figure}[!ht] \centering  
%     \subfigure[Basic A* algorithm planning result]{
%         \label{fig:A_P.sub1}
%         \includegraphics[width=0.47\linewidth,trim={120 0 30 120},clip]{graphics/BasicA2.jpg}}
%     \subfigure[Improved A* algorithm planning result]{
%         \label{fig:A_P.sub2}
%         \includegraphics[width=0.47\linewidth,trim={120 0 30 120},clip]{graphics/ImprovedA2.jpg}}
%     \caption{Comparison of Basic A* Algorithm and Improved A* Algorithm in Parallel Parking Scenario}
%     \label{fig:A_star_parallel}
% \end{figure}

\begin{table*}[!ht]  \small  \centering
\setlength\tabcolsep{2pt} \renewcommand{\arraystretch}{1.0}
    \caption{Algorithm Performance Comparison in Parallel Parking Scenario}
    \begin{tabular}{ccccccc} \toprule
        algorithm & Map loading time & Planning time & Path length & Travel time & Average acceleration & Mean steering Angle \\ \midrule
        Basic A* & 110479.79 & 65.77 & 101.8261 & 26.2 & 1.485 & 12.7033 \\ 
        Improved A* & 3.10 & \textbf{1213.97} & 101.8145 & 66.4 & 0.30 & 20.03\\ \bottomrule
    \end{tabular}
    \label{tab:comparison_parallel}
\end{table*}

In this scenario, the improved A* algorithm achieved over 95\% optimization in terms of time. 
It also demonstrated a 79.8\% improvement in acceleration metrics.
It is worth noting that the traditional A* algorithm failed to plan a path in this scenario, resorting to a combination of numerous abrupt and unrealistic sharp turns and straight lines, which resulted in a seemingly better average turning angle.

This scenario fully demonstrates the limitations of the traditional A* algorithm in local map garage parking tasks. 
The presence of numerous obstacles makes the computational time of the traditional A* algorithm intolerable, and the overly simplistic and rough trajectories completely disregard safety factors and vehicle capabilities.
Due to these shortcomings of the traditional A* algorithm, this paper proposes a targeted improved algorithm.

\subsection{Unreachable Parking Spots}
\label{Unreachable Parking Spots_ans}

The following map depicts a scenario where the parking spot is unreachable, with the starting coordinates at (95,85) and the destination coordinates at (109,117). 
It is important to note that the destination specified by the algorithm is an obstacle, and the car cannot reach the target point.
The destination is marked with a red dot in the figure.

\begin{figure}[ht!]
  \centering
    \begin{subfigure}[t]{0.235\textwidth}
      \centering   
      \includegraphics[width=1\linewidth,trim={30 40 30 35},clip]{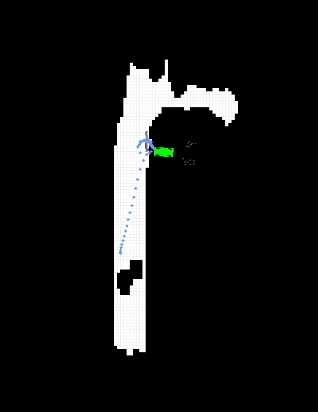}
        \caption{Basic A* algorithm unable to complete planning}
        \label{fig:unreach1}
    \end{subfigure}
    \begin{subfigure}[t]{0.235\textwidth}
      \centering   
      \includegraphics[width=\linewidth,trim={30 40 30 35},clip]{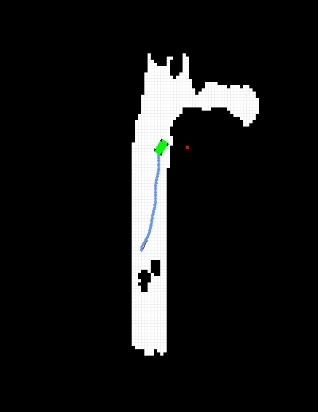}
        \caption{Improved A* algorithm still able to plan}
        \label{fig:unreach2}
    \end{subfigure}
\caption{\label{fig:unreach}Planning results of basic A* algorithm and improved A* algorithm when the parking spot is unreachable (the parking spot is red, located within the obstacle)}
\end{figure}

The A* algorithm is unable to plan a path, thus resulting in anomalous output from the controller. 
Such results are obviously inapplicable to real-world hardware.
For demonstrative purposes, however, the incorrect planning result is still shown in the simulation environment. As shown in Figure \autoref{fig:unreach1}, the path output by the planner (indicated in purple) is clearly disconnected from the starting point, leading to the controller executing a bizarre trajectory (depicted in blue) that cannot satisfy real-world conditions.

The planning result of the improved A* algorithm proposed in this paper is shown in Figure \autoref{fig:unreach2}.
Although the red dot representing the parking destination is within an obstacle and impossible to reach, the improved A* algorithm proposed in this paper still plans a smooth path that is as close to the destination as possible, while avoiding getting too close to the walls.

Experimental validation has shown that the improved algorithm proposed in this paper still has the ability to make attempted planning and strive to provide the best path in special cases where the parking spot is unreachable.

% \begin{figure}[!ht] \centering  
% \subfigure[Basic A* algorithm unable to complete planning]{
% \label{fig:unreach1}
% \includegraphics[width=0.47\linewidth,trim={30 40 30 35},clip]{graphics/unreach.jpg}}\subfigure[Improved A* algorithm still able to plan]{
% \label{fig:unreach2}
% \includegraphics[width=0.47\linewidth,trim={30 40 30 35},clip]{graphics/reach.jpg}}
% \caption{Planning results of basic A* algorithm and improved A* algorithm when the parking spot is unreachable (the parking spot is red, located within the obstacle)}
% \label{fig:unreach}
% \end{figure}

%%%%%%%%%%%%%%%%%%%%%%%%%%%%%%%%%
\subsection{Ablation Experiments}
\label{Ablation Experiments}

\begin{figure}[ht!]
  \centering
    \begin{subfigure}[t]{0.235\textwidth}
      \centering   
      \includegraphics[width=1\linewidth,trim={190 130 160 160},clip]{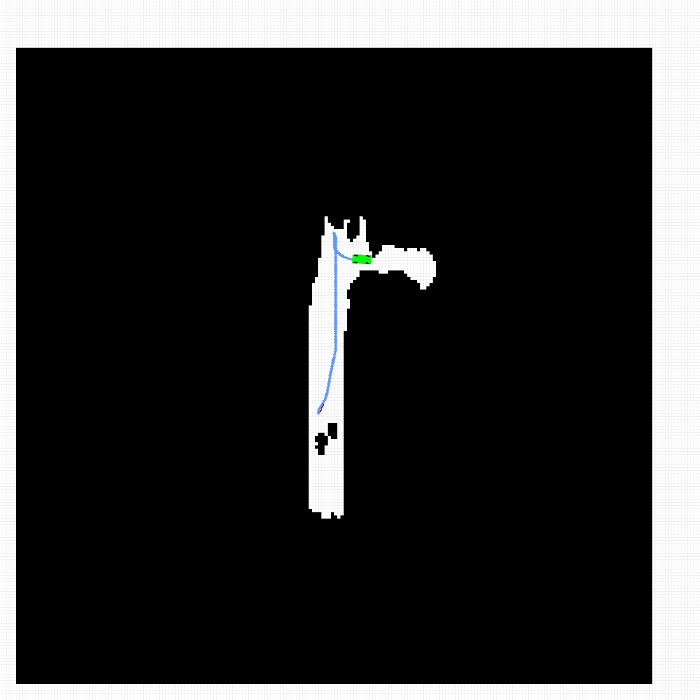}
        \caption{Experiment 1}
        \label{fig:sy1}
    \end{subfigure}
    \begin{subfigure}[t]{0.235\textwidth}
      \centering   
      \includegraphics[width=\linewidth,trim={190 130 160 160},clip]{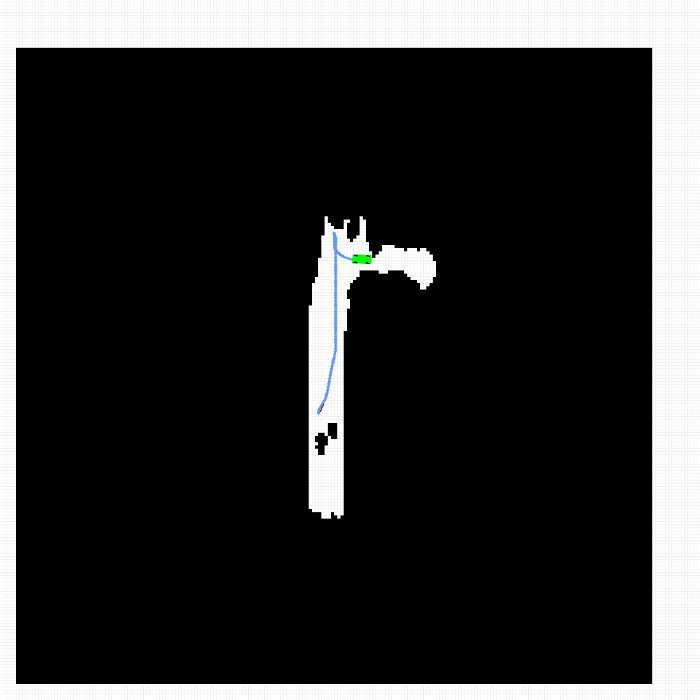}
        \caption{Experiment 2}
        \label{fig:sy2}
    \end{subfigure}
\caption{\label{fig:sy12}Ablation Experiments}
\end{figure}

\begin{figure}[ht!]
  \centering
    \begin{subfigure}[t]{0.235\textwidth}
      \centering   
      \includegraphics[width=1\linewidth,trim={90 90 90 90},clip]{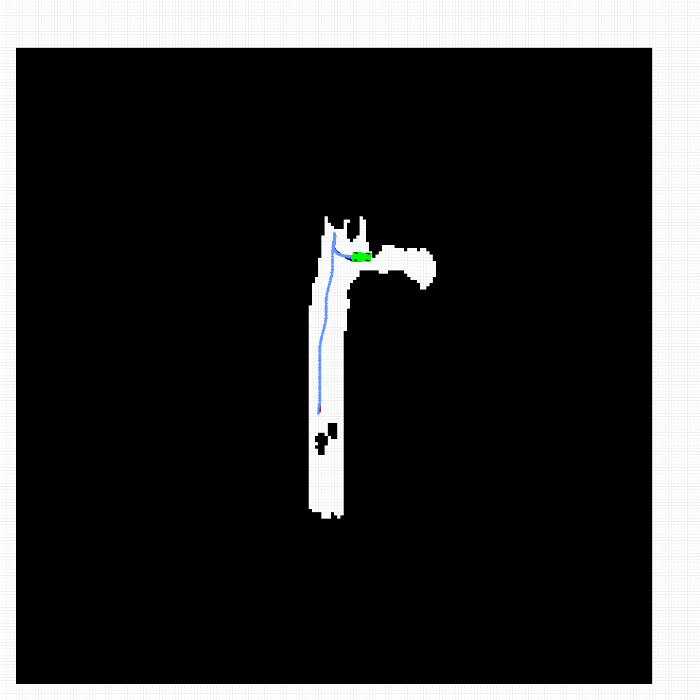}
        \caption{Experiment 3}
        \label{fig:sy3}
    \end{subfigure}
    \begin{subfigure}[t]{0.235\textwidth}
      \centering   
      \includegraphics[width=\linewidth,trim={90 90 90 90},clip]{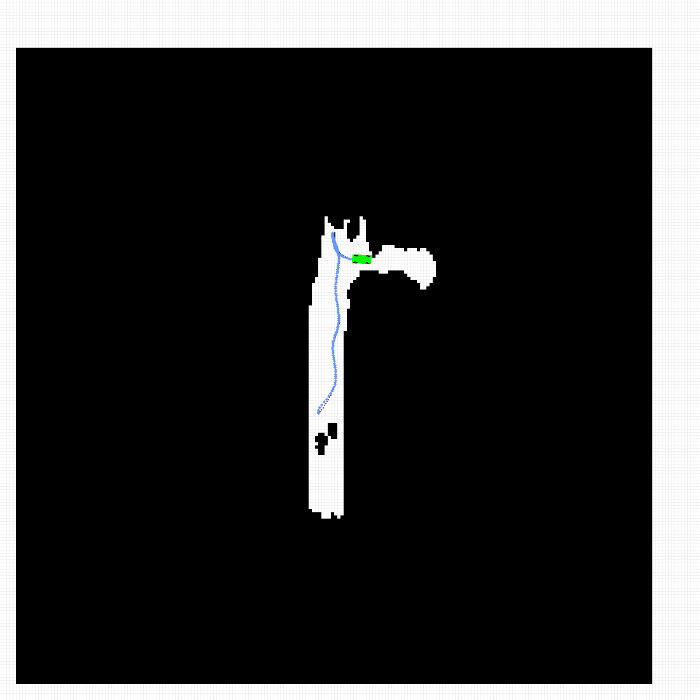}
        \caption{Experiment 4}
        \label{fig:sy4}
    \end{subfigure}
\caption{\label{fig:sy34}Ablation Experiments}
\end{figure}

\begin{figure}[ht!]
  \centering
    \begin{subfigure}[t]{0.235\textwidth}
      \centering   
      \includegraphics[width=1\linewidth,trim={160 135 160 155},clip]{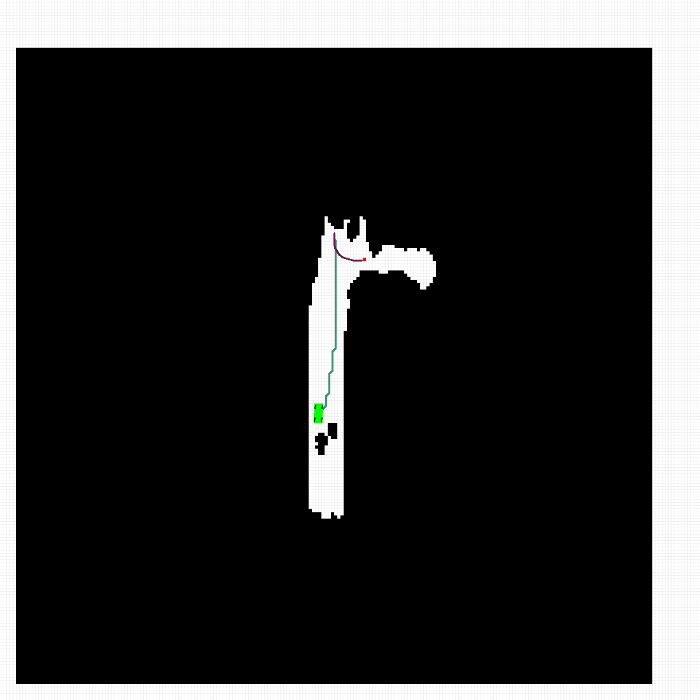}
        \caption{Experiment 5}
        \label{fig:sy5}
    \end{subfigure}
    \begin{subfigure}[t]{0.235\textwidth}
      \centering   
      \includegraphics[width=\linewidth,trim={160 135 160 155},clip]{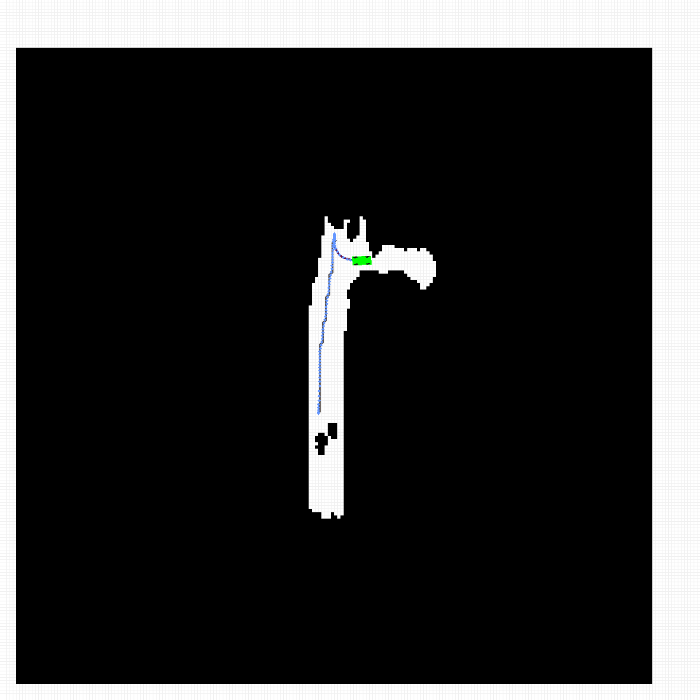}
        \caption{Experiment 6}
        \label{fig:sy6}
    \end{subfigure}
\caption{\label{fig:sy56}Ablation Experiments}
\end{figure}

The ablation study planning results are shown in \autoref{fig:sy12}, \autoref{fig:sy34}, and \autoref{fig:sy56}.  In the ablation experiments, the maps, start positions, end positions, and parking categories used in each group of experiments were consistent with those used in \autoref{Vertical Parking}.

The results of Experiment 1 and Experiment 2 indicate that not using the optimizations of heuristic weighting and dynamic obstacle loading do not necessarily affect the final generated path, but subsequent data can prove that not using these optimizations can have a negative impact on the algorithm's planning speed.

% \begin{figure}[!ht] \centering  
% \subfigure[Experiment 1]{
% \label{fig:sy1}
% \includegraphics[width=0.47\linewidth,trim={190 130 160 160},clip]{graphics/sy1.png}}\subfigure[Experiment 2]{
% \label{fig:sy2}
% \includegraphics[width=0.47\linewidth,trim={190 130 160 160},clip]{graphics/sy2.png}}
% \caption{Ablation Experiments}
% \label{fig:sy12}
% \end{figure}

The planning results of Experiment 3 and Experiment 4 are shown in \autoref{fig:sy34}.
It can be seen that not using bidirectional A* and binary heap optimizations can lead to unnecessary turns in the path, and measurement data shows that abandoning these optimizations can also reduce the search efficiency of the algorithm. 
However, Experiment 4, which enables neighborhood expansion, exhibits issues with low path quality, resulting in significant unnecessary bends.
A possible reason is that neighborhood expansion can lead to excessively long single-step distances, making it difficult for the current nodes of the bidirectional search to connect smoothly, resulting in reduced path quality.

% \begin{figure}[!ht] \centering  
% \subfigure[Experiment 3]{
% \label{fig:sy3}
% \includegraphics[width=0.47\linewidth,trim={90 90 90 90},clip]{graphics/sy3.png}}\subfigure[Experiment 4]{
% \label{fig:sy4}
% \includegraphics[width=0.47\linewidth,trim={90 90 90 90},clip]{graphics/sy4.png}}
% \caption{Ablation Experiments}
% \label{fig:sy34}
% \end{figure}

The paths generated by Experiment 5 and Experiment 6 are generally similar to those generated by the improved A* algorithm and Experiment 3, indicating that the Bezier curve optimization does not significantly affect the general shape of the path. 
However, it is evident that paths without Bezier curve optimization exhibit a large number of sharp corners, which is an inherent disadvantage of the grid-based graph search algorithm such as the basic A*.  
Therefore, the results of Experiment 5 and Experiment 6 further demonstrate the necessity of interpolation optimization.

% \begin{figure}[!ht] \centering  
% \subfigure[Experiment 5]{
% \label{fig:sy5}
% \includegraphics[width=0.47\linewidth,trim={160 135 160 155},clip]{graphics/sy5.png}}\subfigure[Experiment 6]{
% \label{fig:sy6}
% \includegraphics[width=0.47\linewidth,trim={160 135 160 155},clip]{graphics/sy6.png}}
% \caption{Ablation Experiments}
% \label{fig:sy56}
% \end{figure}

Based on the experimental results, the experimental parameters concerned in this study are recorded.  
A data comparison is shown in \autoref{tab:Ablation Experiment}:

\begin{table*}[!ht]  \centering \small
    \setlength\tabcolsep{2pt} \renewcommand{\arraystretch}{1.0}
    \caption{Comparison of each algorithm in ablation experiment}
    \begin{tabular}{lcccc}  \toprule
        Algorithm & Map Loading Time & Planning Time & Average Acceleration & Average Steering Angle \\ \midrule
        Basic A* & 98255.68 & 21.86 & 1.812 & 19.74  \\ 
        Experiment 1 & 1.78 & 442.93 & 0.306 & 10.79  \\ 
        Experiment 2 & 96384.52 & 6.9 & 0.3062 & 10.79  \\ 
        Experiment 3 & 1.95 & 609.01 & 0.141 & 12.89  \\ 
        Experiment 4 & 1.85 & 583.30 & 0.631 & 20.31  \\ 
        Experiment 5 & 1.82 & 430.61 & 1.662 & 10.05  \\ 
        Experiment 6 & 1.80 & 621.65 & 1.555 & 9.20  \\ 
        Improved A* & 1.81 & 411.04 & 0.306 & 10.79 \\ \bottomrule
    \end{tabular}
    \label{tab:Ablation Experiment}
\end{table*}
 
%%%%%%%%%%%%%%%%%%%%%%%%%%%%%%%%%%%%%%%%%%%%%%
\section{Discussion}
\label{sec:discussion}
%%%%%%%%%%%%%%%%%%%%%%%%%%%%%%%%%%%%%%%%%%%%%%

The study conducts a series of experiments in diverse scenarios to compare the computational efficiency of the improved A* algorithm and the traditional A* algorithm for generating navigation paths. 
The findings demonstrate that, considering the volume of the ego vehicle, the proposed improved A* algorithm generally achieved a computational speed increase of 95\% or more, with particularly significant enhancements in map loading time and planning time. 
This validates that the performance optimization methods outlined in this paper are both feasible and effective, addressing issues such as excessive search nodes and prolonged single search times in the traditional A* algorithm, thereby significantly enhancing its planning time efficiency.

In terms of comfort validation, this study comprehensively evaluates the overall planning process proposed by collecting vehicle motion parameters returned by the controller and comparing them with a traditional method combining basic A* with geometric curve method.
The data illustrate that paths generated by our proposed method result in smooth vehicle movement without frequent speed and direction changes, leading to more decisive vehicle behavior.

The proposed method showcased better real-time performance and safety metrics crucial for autonomous parking tasks while also offering higher comfort compared to traditional methods.

The results of Experiment 1 indicate that heuristic weighting can enhance computational efficiency to a certain extent.
However, in the case of multiple optimization methods working simultaneously, the presence or absence of heuristic weighting has a minor impact on algorithm efficiency.

The results of Experiment 2 demonstrate that the dynamic obstacle loading optimization significantly improves the overall efficiency of the algorithm. 
Other optimizations applied to the planning itself also lead to an enhancement in the performance of the planning algorithm. 
Compared to the baseline A*, the planning part excluding map reading reduced the runtime by 68.4\%.

The findings of Experiment 3 reveal that bidirectional search and binary heap optimization enhance the planning efficiency of the algorithm.
Compared to Experiment 3, the modified A* method applying these optimizations exhibits a 32.5\% improvement in computational efficiency. 
This efficiency gain is attributed to the bidirectional search, where both current nodes aim to quickly approach the target node in the first half of the search, thereby reducing search time. 
The excellent performance of the binary heap further boosts algorithm efficiency through its data structure.

The results of Experiment 4 suggest that neighborhood expansion is not suitable for all scenarios and should not be used indiscriminately.
Therefore, the final improved A* algorithm adopted in this paper does not include the neighborhood expansion strategy, but its functionality is preserved.

Experiment 5 confirms that Bezier curve optimization can indeed significantly enhance the smoothness of the algorithm-generated trajectory while barely affecting time efficiency.

Experiment 6 further validates the necessity of bidirectional search and binary heap optimization.

In summary, the multi-dimensional optimization method proposed in this design can effectively enhance the performance of the A* planning algorithm.
In practical applications, suitable combinations of optimization methods can be selected based on specific requirements.

%%%%%%%%%%%%%%%%%%%%%%%%%%%%%
\subsection{Limitations}

Although this paper has achieved some results in improving the A* algorithm, there are still some deficiencies and areas for improvement.
For example, the algorithm cannot handle non-standard parking spaces, nor can it handle cases where the area near the parking space is too narrow, making it impossible to park by combining straight lines and arcs. 
Additionally, real-world conditions may not always satisfy all the assumptions of the model mentioned in \autoref{Vehicle Kinematic Model}, and this paper has not yet considered dynamic constraints such as tire slip that may affect vehicle movement. 
In future work, this research will continue to explore improved methods for the A* algorithm and attempt to combine it with other advanced technologies to further enhance the performance and stability of the automatic parking system. 
For instance, more complex multi-segment trajectories can be employed to achieve parking logic, addressing more intricate parking space conditions.

%%%%%%%%%%%%%%%%%%%%%%%%%%%%%%%%%%%%%%%%%%%%%%
\section{Conclusions}
\label{sec:conclusion}
%%%%%%%%%%%%%%%%%%%%%%%%%%%%%%%%%%%%%%%%%%%%%%

This paper proposes targeted improvements to the A* algorithm for the automatic parking task and introduces a series of optimization measures, aiming to enhance the algorithm's performance and applicability. 
Through a series of experimental verifications, the proposed improved method has achieved remarkable results, significantly optimizing the trajectory generation quality and planning time consumption of the A* algorithm.

Firstly, aiming at the issue of low efficiency in the search process of the A* algorithm, this method designs a series of measures to enhance performance. 
By optimizing the heuristic function, the node selection strategy is improved, reducing unnecessary searches and thus enhancing the search speed of the algorithm. 
Simultaneously, a bidirectional search method is adopted to reduce the search space of the algorithm. During the search execution, the data structure implemented by the optimized algorithm is improved, allowing the algorithm to run faster. 
These improvements enable the algorithm to find feasible paths more quickly when dealing with complex parking scenarios.

Secondly, considering the characteristics of the automatic parking task, this method optimizes the trajectory quality generated by the planning algorithm using neighborhood expansion and Bezier curve interpolation, enabling the algorithm to generate smoother and more continuous paths. 
At the same time, the MPC control algorithm is improved, making the output control instructions focus more on the safety and comfort of parking. 
This improvement enhances the planning and control precision of the algorithm, significantly improving the trajectory quality.
It also realizes attempted planning for unreachable parking spaces, enabling the algorithm to handle some special situations.

This paper studies and improves the application of the A* algorithm in automatic parking tasks, achieving meaningful results. 
% With continuous technological development and improvement, automatic parking planning and control technology will continue to progress, and automatic parking systems will be more widely applied and promoted in the future.
In the future, we will apply knowledge graph \cite{lan2022semantic,liu2022towards,liu2024mining} to improve this work.
\bibliographystyle{IEEEtran}
\bibliography{bibliography}

\end{document}